\documentclass{svjour3}
\journalname{Machine learning}

\usepackage[hidelinks]{hyperref}

\newif\ifcomments

\usepackage{subcaption}
\usepackage{booktabs}
\usepackage{listings}
\usepackage{pgfplots}
\usepackage{tikz}
\usepackage{algorithm2e}
\usepackage{inconsolata}
\usepackage{standalone}
\usepackage{xr}
\usepackage{stfloats}
\usepackage{adjustbox}
\usepackage{hyperref}
\usepackage{ifdraft}
\usepackage{amsmath}
\usepackage[misc,geometry]{ifsym}

\newtheorem{prop}{Proposition}

\newcommand{\tw}[1]{\texttt{#1}}

\newcommand{\popper}{\textsc{Popper}}
\newcommand{\metagol}{\textsc{Metagol}}
\newcommand{\ilasp}{\textsc{ILASP}}
\newcommand{\alephilp}{\textsc{Aleph}}
\newcommand{\name}{\textsc{Hempel}}
\newcommand{\nameold}{\popper}
\usepackage{amssymb,cancel}
\newcommand{\metagoln}{
\textsc{Metagol}$_{\scalebox{0.6}{\cancel{\rotatebox[origin=c]{245}{$\circlearrowleft$}}}}$
}

\usepackage{xcolor}
\usepackage[normalem]{ulem} 

\usepackage{xargs}

\ifcomments{
\paperwidth=\dimexpr \paperwidth + 12cm\relax
\oddsidemargin=\dimexpr \oddsidemargin + 6cm\relax
\evensidemargin=\dimexpr \evensidemargin + 6cm\relax
\marginparwidth=\dimexpr \marginparwidth + 5cm\relax
\usepackage[colorinlistoftodos,prependcaption]{todonotes}
\usepackage{etoolbox}

\newcommandx{\ac}[3][1=,3=]{%
\ifblank{#1#2}{}{%
\todo[linecolor=red,backgroundcolor=white,bordercolor=red]{%
\ifblank{#1}{}{\color{red}\sout{#1}\\}%
\textcolor{red}{
\Large
#2%
}%
}%
}%
\ifblank{#3}{}{{\color{red}#3}}%
}

\newcommandx{\rolf}[3][1=,3=]{%
\ifblank{#1#2}{}{%
\todo[linecolor=blue,backgroundcolor=white,bordercolor=blue]{%
\ifblank{#1}{}{\color{blue}\sout{#1}\\}%
\textcolor{blue}{
\Large
#2%
}%
}%
}%
\ifblank{#3}{}{{\color{blue}#3}}%
}
\else 
\newcommandx{\ac}[3][1=,3=]{#3}
\newcommandx{\rolf}[3][1=,3=]{#3}
\fi 

\lstset{
    mathescape=true,
    basicstyle=\ttfamily,
    captionpos=b,
    escapeinside={(*}{*)},
    columns=flexible
}

\lstnewenvironment{myalgorithm}[1][] 
{
    \lstset{ 
	basicstyle=\ttfamily,
        mathescape=true,
        numbers=left,
        escapeinside={@}{@},
        columns=flexible,
        numberstyle=\ttfamily,
        keywordstyle=\bfseries\ttfamily,
        keywords={,and, return, not, def, in, if, else, for, foreach, while, }
        numbers=left,
        xleftmargin=.02\textwidth,
    }
    \footnotesize
}
{}

\usepackage{pgfkeys}
\newenvironment{customlegend}[1][]{%
\begingroup
\csname pgfplots@init@cleared@structures\endcsname
\pgfplotsset{#1}%
}{%
\csname pgfplots@createlegend\endcsname
\endgroup
}%
\def\addlegendimage{\csname pgfplots@addlegendimage\endcsname}

\title{Learning logic programs by explaining their failures}

\author{Rolf Morel \and Andrew Cropper}

\institute{
R. Morel \Letter \at
University of Oxford\\
\email{rolf.morel@cs.ox.ac.uk}\\
\and
A. Cropper\at
University of Oxford\\
\email{andrew.cropper@cs.ox.ac.uk}
}

\begin{document}

\maketitle

\abstract{
Scientists form hypotheses and experimentally test them.
If a hypothesis fails (is refuted), scientists try to \emph{explain} the failure to eliminate other hypotheses.
The more precise the failure analysis the more hypotheses can be eliminated.
Thus inspired, we introduce failure explanation techniques for inductive logic programming.
Given a hypothesis represented as a logic program, we test it on examples.
If a hypothesis fails, we explain the failure in terms of failing sub-programs.
In case a positive example fails, we identify failing sub-programs at the granularity of literals.
We introduce a failure explanation algorithm based on analysing branches of SLD-trees.
We integrate a meta-interpreter based implementation of this algorithm with the test-stage of the \nameold{} ILP system.
We show that fine-grained failure analysis allows for learning fine-grained constraints on the hypothesis space.
Our experimental results show that explaining failures can drastically reduce hypothesis space exploration and learning times.
}


\section{Introduction}
\label{sec:introduction}

Explanations are ubiquitous in our cognitive lives \cite{keil2000explanation}.
They are crucial to the process of forming hypotheses, testing them on data, analysing the results, and forming new hypotheses, that is to say, to science \cite{popper2002conjectures}.
For instance, imagine Alice is a chemist trying to synthesise a vial of a compound from two substances (e.g.~\emph{synth(thaum,slood,octiron)}).
Alice can perform actions, such as fill a vial with a substance (\emph{fill(Vial,Sub)}) or mix two vials (\emph{mix(V1,V2,V3)}), and sequence them to form a hypothesis, e.g.:

\begin{center}
\emph{synth(A,B,C) $\leftarrow$ fill(V1,A), fill(V1,B), mix(V1,V1,C)}
\end{center}

\noindent
This hypothesis says that to synthesise a vial of compound \emph{C}, fill vial \emph{V1} with substance \emph{A}, fill vial \emph{V1} with substance \emph{B}, and mix vial \emph{V1} with itself to form \emph{C}.

When Alice experimentally tests this hypothesis she finds that it \emph{fails}.
From this failure Alice concludes (\textbf{C1}) that hypotheses which add further actions (i.e.~literals) will also fail.
However, as Alice observed that the second action caused the failure, she can \emph{explain} the failure as ``vial \emph{V1} cannot be filled a second time''.
This allows her to conclude (\textbf{C2}) that any hypothesis that includes \emph{fill(V1,A)} and \emph{fill(V1,B)} will fail.
Clearly, conclusion \textbf{C2} allows Alice to eliminate more hypotheses than \textbf{C1}.
That is, by explaining failures Alice can better form new hypotheses.

We formalise this mode of reasoning for explaining failures of logical theories.
We do so in the context of inductive program synthesis, where the goal is to machine learn computer programs from data \cite{mis}.
Existing inductive logic programming (ILP) approaches fail to generalise from observed failures.
Many ILP systems \cite{atom,ilasp3,popper} only learn from the failure of an entire hypothesis -- as Alice does when she concludes \textbf{C1} -- and cannot explain why a hypothesis fails, e.g.~cannot reason like Alice does to conclude \textbf{C2}.
Some systems can identify parts of a program that cause a failure, but cannot learn from this information.
For instance, Metagol \cite{metagol} will repeatedly retry failing program fragments.

We address these limitations by automatically explaining program failures, taking inspiration from algorithmic debugging \cite{ADsurvey}.
The idea is to analyse the failure of a hypothesis to identify \emph{sub-programs} that also fail.
To illustrate, consider hypothesis \emph{H$_1$}:
\[
\left\{
\begin{array}{l}
    \emph{droplast(A,B) $\leftarrow$ empty(A),tail(A,B)}
\end{array}
\right\}
\]
\noindent
If \emph{droplast([1,2],[1])} is a positive example, then \emph{H$_1$} does not cover this example.
From this failure we can learn that \emph{H$_1$}'s sub-program \{ \emph{droplast(A,B) $\leftarrow$ empty(A)} \} also does not cover this example.
We show that by identifying failing sub-programs and accumulating constraints generated from them, we can eliminate more hypotheses (e.g.~any single clause program that expands the above sub-program).
When the overhead of failure explanation is low, our approach reduces learning times.

Most logic program debugging systems \cite{KohlerDatalogDebugging,profl} and some synthesis systems \cite{mis,prosynth} can identify a subset of clauses as being the cause of a failure.
We additionally identify literals \emph{within} clauses responsible for failure (without the requirement of trace-complete examples needed by theory revision systems such as FORTE \cite{forte}).
We show that this fine-grained failure analysis allows for learning finer-grained constraints on the hypothesis space.

Our contributions are:
\begin{itemize}
\item We relate logic programs that fail on examples to their failing sub-programs.
For wrong answers we identify clauses.
For missing answers we additionally identify literals within clauses.
\item We show that hypotheses that are specialisations and generalisations of failing sub-programs can be eliminated,
and prove that hypothesis space pruning based on sub-programs is more effective than pruning without them.
\item We introduce \name{}, an ILP system extending the \nameold{} ILP system, which analyses SLD-trees to automatically explain failures in terms of sub-programs.
\item We experimentally show that failure explanation can drastically reduce (i) hypothesis space exploration and (ii) learning times.
\end{itemize}

\section{Related work}
\label{sec:related}

\paragraph{\textbf{Program synthesis.}}
Inductive program synthesis systems automatically generate computer programs from specifications, typically input/output examples \cite{mis}.
This topic interests researchers from many areas of machine learning, including Bayesian inference \cite{silver:aaai20} and neural networks \cite{dreamcoder}.
We focus on ILP techniques, which induce logic programs \cite{mugg:ilp}.

\paragraph{\textbf{Recursion}.}
Both classical ILP systems \cite{progol,tilde,aleph} as well as many modern ones, e.g.~Atom \cite{atom}, struggle to learn recursive programs, or cannot learn them at all, e.g. Inspire \cite{inspire} and ~FastLAS \cite{fastlas}.
By contrast, our system, \name{}, can learn recursive programs and thus programs that generalise to input sizes it was not trained on.
Compared to many modern ILP systems \cite{dilp,hexmil,apperception}, \name{} supports large and infinite domains, which is important when reasoning about complex data structures, such as lists.
In addition, unlike many state-of-the-art systems \cite{metagol,dilp,hexmil,celine:bottom}, \name{} does not require metarules (i.e.~program templates) to restrict the hypothesis space.

\paragraph{\textbf{Algorithmic debugging}.}
Algorithmic debugging \cite{ADsurvey} explains failures in terms of sub-programs.
Alongside his seminal work on logic program synthesis, Shapiro \cite{mis} introduced the notion of \emph{debugging trees} for semi-automated identification of failing clauses.
Only being able to return clauses responsible for entailing an atom is still the standard for logic programming debugging \cite{KohlerDatalogDebugging,profl}.
Unlike these systems, we automatically identify literals within clauses which cause an atom to not be entailed, and integrate the failure explanation process in a program synthesis system.

\paragraph{\textbf{Theory revision and repair}.}
Shapiro's Model Inference System (MIS) \cite{mis} is a theory revision system which, through interaction with a user, is capable of synthesising programs.
MIS uses SLD-trees to determine which clauses of a program are responsible for entailing a negative example, at which point the user needs to say which of these clauses is wrong.
To cover a non-covered positive example, additional clauses get added, possibly involving user-interaction, without regard for why the current clauses do not entail this example.
By contrast, \name{} does not require an oracle and can automatically identify clauses and literals within clauses as being responsible for not entailing a positive example.

There are theory revision systems \cite{wrobel1996fotr} able to identify literals as \emph{revision points} within theories, though often with limitations.
Some require user-interaction \cite{clint,focl}.
FORTE \cite{forte} uses hill-climbing to gradually revise a theory, heuristically following revisions that improve training accuracy.
Unlike FORTE, \name{} is guaranteed to find an optimal solution if one exists.
FORTE can automatically identify responsible literals of a sub-program, given that the examples are trace-complete, i.e.~all necessary recursive calls of the target predicate are included as positive examples.
Our failure explanation algorithm automatically identifies responsible clauses and literals which cause a program to not entail an atom, without any condition on the examples.

In general, theory revision and theory repair \cite{bundy2016reformation} are concerned with updating a current hypothesis by applying generalisation and specialisation operators to the identified revision points.
Whereas these systems refine a single program at a time, \name{} uses the failure of a (sub-)program to \emph{refine the hypothesis space}, each time pruning away a large class of programs.

\paragraph{\textbf{Failure explanation}.}
Some modern ILP systems can be said to have a degree of failure explanation.

\metagol{} \cite{metagol} is a meta-interpreter which uses examples to drive the search, gradually building up a program whilst partially evaluating it on an example.
When a failure occurs, Metagol knows it is due to the last literal that was added, which causes it to backtrack.
However, due to its iterative deepening strategy, \metagol{} will reconsider these program fragments many times, and has no way to learn from failures.
By contrast, \name{} learns constraints which ensure that failing program fragments are never reconsidered.

\ilasp3{} \cite{ilasp3} learns recursive ASP programs, with \emph{partial interpretations} serving as examples.
It starts by enumerating the space of candidate rules, assigning each an id.
Next a select-test-constrain loop selects a hypothesis, a subset of the candidate clauses, based solely on constraints over the ids.
When a model of a selected hypothesis does not correctly extend the given partial interpretations, the hypothesis fails with the model being its \emph{violating reason}.
Constraints can be derived from a violating reason by checking which combinations of candidate rules also have it as a model, which is an expensive operation.
\name{}'s learning of constraints by identifying sub-programs is more efficient and, by defining its hypothesis selection problem over literals, it is not restricted to identifying just clauses as causing a failure.

Like \ilasp3{}, \textsc{ProSynth} \cite{prosynth} precomputes every possible clause and employs a select-test-constrain loop over clause ids.
\textsc{ProSynth} uses the notion of \emph{query provenance} \cite{CheneyProvenance} for identifying which clauses of a hypothesis are responsible for (not) entailing an example, encoding identified subsets as constraints.
\textsc{ProSynth} learns Datalog programs, which is just a fragment of the definite programs which can be learned by \name{}.
Additionally, \name{}'s failure explanation is finer grained as it also identifies which literals cause failure.

\paragraph{\textbf{Learning from failures.}}
Our system builds on \popper{} \cite{popper}, see Section \ref{sec:implementation}.
\popper{} learns first-order constraints by a process that is similar to conflict-driven clause learning \cite{CDCLhandbook}.
The constraints that Popper learns are always based on entire hypotheses (i.e.~it only reasons as Alice does for conclusion \textbf{C1} in the introduction).
\ac{you could mention that it uses subsumption to reason about the generality of failures}
\name{}'s failure explanation can hence be viewed as allowing \popper{} to detect smaller, finer-grained conflicts, yielding smaller and more general constraints which prune more effectively (which brings the reasoning about failures up to the level of conclusion \textbf{C2}).

\section{Problem setting}
\label{sec:framework}

In this section, we (i) describe our problem setting; (ii) relate specialisations and generalisations to missing and incorrect answers; (iii) define failing sub-programs; and (iv) show that sub-programs lead to better pruning.


\paragraph{Preliminaries.}
We assume standard logic programming definitions \cite{lloyd:book}.
We define $\theta$-subsumption \cite{plotkin:thesis,midelfart}.
A clause $C_1$ \emph{subsumes} a clause $C_2$ iff there exists a substitution $\theta$ such that $C_1\theta \subseteq C_2$.
A clausal theory $T_1$ subsumes a clausal theory $T_2$ 
iff $\forall C_2 \in T_2, \exists C_1 \in T_1$ such that $C_1$ subsumes $C_2$.
Subsumption implies entailment, i.e. if $T_1$ subsumes $T_2$ then $T_1 \models T_2$.




\subsection{Learning from failures}

We adopt the learning from failures (LFF) approach to ILP \cite{popper}.
Let $\mathcal{H}$ be a set of hypotheses, where each hypothesis is a definite program (a set of definite clauses).
Hypothesis space pruning is made explicit in LFF by means of \emph{hypothesis constraints}.
For our purposes, it suffices to see a hypothesis constraint as a set of programs, typically related by their syntax, where the purpose of this set is to \emph{prune}, i.e.~rule out, these hypotheses.
For example, given a program $P$, a hypothesis constraint could prune any program $Q \in \mathcal{H}$ such that $P \subseteq Q$, i.e.~any program that adds clauses to $P$.
Given a set of hypothesis constraints $C = \{ C_1, \ldots, C_n \}$, $\mathcal{H}_C = \mathcal{H} \setminus (C_1 \cup \ldots \cup C_n)$ denotes the set of all hypotheses \emph{not} pruned by the individual constraints.

We define LFF's input\footnote{We work with a more abstract LFF input than its original definition: our hypothesis spaces and its constraints are just sets rather than sets being represented by formulae in a constraint satisfaction language.} and introduce our running example:



\begin{definition}[\textbf{LFF input}]
\label{def:probin}
A \emph{LFF} input is a tuple $(E^+, E^-, \mathcal{H}, B, C)$ where
$E^+$ and $E^-$ are sets of ground atoms denoting positive and negative examples respectively; $\mathcal{H}$ is a set of hypotheses;
$B$ is a definite program denoting background knowledge\footnote{The background knowledge program can make use of functional symbols.};
and $C$ is a set of hypothesis constraints.
\end{definition}

\begin{example}
\label{ex:lff_input}
To illustrate LFF, consider an input for learning a \emph{droplast/2} program.
Suppose our hypotheses $\mathcal{H}$ are definite programs with \emph{droplast/2} in the head of each clause and \emph{droplast/2}, \emph{empty/1}, \emph{head/2}, \emph{tail/2} and \emph{cons/3} occurring in bodies.
Our background knowledge $B$ consists of definitions for these predicates, except for \emph{droplast/2}.
$E^+ = \{ \emph{droplast}([1,2,3],[1,2]), \emph{droplast}([1,2],[1]) \}$ and $E^- = \{ \emph{droplast}([1,2],[]) \}$ are our positive and negative examples.
Our set of hypothesis constraints $C$ is initially empty.
\end{example}
\noindent
We define a LFF solution:

\begin{definition}[\textbf{LFF solution}]
\label{def:solution}
Given an input tuple $(E^+, E^-, \mathcal{H}, B, C)$, a hypothesis $H \in \mathcal{H}_C$ is a \emph{solution} when $H$ is \emph{complete} ($\forall e \in E^+, \; B \cup H \models e$) and \emph{consistent} ($\forall e \in E^-, \; B \cup H \not\models e$).
\end{definition}

\noindent
If a hypothesis is not a solution then it is a \emph{failing} hypothesis.
A hypothesis $H$ is \emph{incomplete} when $\exists e^+ \in E^+, \; H \cup B \not \models e^+$.
A hypothesis $H$ is \emph{inconsistent} when $\exists e^- \in E^-, \; H \cup B \models e^-$.
A hypothesis $H_1$ is a \emph{specialisation} of hypothesis $H_2$ when $H_2$ subsumes $H_1$. 
Symmetrically, a hypothesis $H_1$ is a \emph{generalisation} of hypothesis $H_2$ when $H_1$ subsumes $H_2$. 

Key to LFF is the ability to learn hypothesis constraints from failed hypotheses.
Given an incomplete hypothesis $H$, a \emph{specialisation constraint} prunes specialisations of $H$. 
Similarly, given an inconsistent hypothesis $H'$, a \emph{generalisation constraint} prunes generalisations of $H'$. 
These constraints are \emph{sound}, that is, they do not prune solutions.


\subsection{Missing and incorrect answers}

Given background knowledge $B$, the failure of a hypothesis $H$ is due to at least one example.
We adopt the following terminology from the algorithmic debugging community \cite{mis,ADsurvey}.
A positive example $e^+$ is a \emph{missing answer} when $B \cup H \not\models e^+$.
Similarly, a negative example $e^-$ is an \emph{incorrect answer} when $B \cup H \models e^-$.
We relate missing and incorrect answers to specialisations and generalisations.
If $H$ has a missing answer $e^+$, then, as a specialisation $H'$ of $H$ entails at most as much as $H$, $e^+$ is a missing answer of $H'$ as well.
Hence all specialisations of $H$ are incomplete and can be eliminated.
Similarly, as generalisations of $H$ entail at least as much as $H$, if $e^-$ is an incorrect answer of $H$, all generalisations of $H$ are inconsistent and can be pruned.



\begin{example}[\textbf{Missing answers and specialisations}]
Given the LFF input from Example \ref{ex:lff_input}, consider the following \emph{droplast} hypothesis:%
\[
\emph{H$_1$} = \left\{
\begin{array}{l}
    \emph{droplast(A,B) $\leftarrow$ empty(A),tail(A,B)}
\end{array}
\right\}
\]
Both $\emph{droplast}([1,2,3],[1,2])$ and $\emph{droplast}([1,2],[1])$ are missing answers of $H_1$, so $H_1$ is incomplete and we can prune its specialisations, e.g.~programs that add literals to the clause.
\end{example}

\begin{example}[\textbf{Incorrect answers and generalisations}]
Consider hypothesis $H_2$:%
\[
\emph{H$_2$} = \left\{
\begin{array}{l}
    \emph{droplast(A,B) $\leftarrow$ tail(A,C),tail(C,B)} \\
    \emph{droplast(A,B) $\leftarrow$ tail(A,B)}
\end{array}
\right\}
\]

\noindent
In addition to being incomplete, $H_2$ is inconsistent because of the incorrect answer $\emph{droplast}([1,2],[])$, so along with specialisations we can prune the generalisations of $H_2$, e.g.~programs with additional clauses.
\end{example}

%
%
%
%

\subsection{Failing sub-programs}

\label{sec:answerstoconstraints}
We now consider explaining failures in terms of failing sub-programs.
The idea is to identify sub-programs that cause the failure.
Consider the following two examples:


\begin{example}[\textbf{Explain missing answer}]
\label{ex:explain_failures}
Consider previously defined $H_1$ and positive example $e^+ = \emph{droplast}([1,2],[1])$.
An explanation for why $H_1$ does not entail $e^+$ is that $\emph{empty}([1,2])$ fails.
It follows that $e^+$ is a missing answer of
$
\emph{H$_1'$} = \left\{
\begin{array}{l}
    \emph{droplast(A,B) $\leftarrow$ empty(A)}\\
\end{array}
\right\}
$.
As $H_1'$ is incomplete we can prune all of its specialisations.
\end{example}


\begin{example}[\textbf{Explain incorrect answer}]
Consider negative example $e^- = \emph{droplast}([1,2],[])$ and $H_2$.
The first clause of $H_2$ always entails $e^-$ irrespective of other clauses being part of the hypothesis.
It follows that $e^-$ is an incorrect answer of $
\emph{H$_2'$} = \left\{
\begin{array}{l}
\emph{droplast(A,B) $\leftarrow$ tail(A,C),tail(C,B)}\\
\end{array}
\right\}
$.
As $H_2'$ is inconsistent we can prune all of its generalisations.

\end{example}

\noindent
Note that when a system like \popper{} observes that $H_2$ fails, it is not able to prune based on $H_2'$.
Whilst costly, an ILP system like ProSynth could learn that $H_2'$ fails.
Given $H_1$ and its failure, \popper{}, \ilasp3{} and ProSynth are unable to determine it is possible to prune based on $H_1'$.

We now define a \emph{sub-program}:

\begin{definition}[\textbf{Sub-program}]
A definite program $P$ is a \emph{sub-program} of a definite program $Q$ if and only if either:
\begin{itemize}
\setlength\itemsep{0pt}
\setlength\parskip{0pt}
\item $P$ is the empty set
\item there exists clauses $C_p \in P$ and $C_q \in Q$ such that $C_p \subseteq C_q$ and $P \setminus \{ C_p \}$ is a sub-program of $Q \setminus \{C_q\}$
\end{itemize}
\end{definition}

\noindent
In this definition, arguments of literals must be syntactically the same\footnote{Our definition hence insists on variable names in literals of a sub-program $Q$ being the same as variable names in the corresponding literals of program $P$.} for the clause subset check to succeed.
In functional program synthesis, sub-programs are typically defined by leaving out nodes in the parse tree of the original program (e.g., \cite{feng:neo}).
Our definition generalises this idea by allowing for arbitrary ordering of clauses and literals. 

In the above examples, $H_1'$ is a sub-program of $H_1$ and so is $H_2'$ of $H_2$.
Note that clauses and literals can be dropped at the same time, e.g.~$
\left\{
\begin{array}{l}
\emph{droplast(A,B) $\leftarrow$ tail(A,C)}\\
\end{array}
\right\}
$ is another sub-program of $H_2$.

We define the failing sub-programs problem:
\begin{definition}[\textbf{Failing sub-programs}]
\label{def:subprogprob}
Given definite program $P$ and sets of examples $E^+$ and $E^-$, the \emph{failing sub-programs problem} is to find all sub-programs of $P$ that do not entail an example of $E^+$ or do entail an example of $E^-$.
\end{definition}

\noindent
By definition, a failing sub-program has a missing answer and/or an incorrect answer.
Hence we can always prune specialisations and/or generalisations of a failing sub-program.
We show that sub-programs are effective at pruning:


\begin{theorem}[\textbf{Better pruning}]
\label{prop:pruning}
Let $H$ be a definite program that fails and $P$ ($\neq H$) be a sub-program of $H$ that fails.
Let $C(H)$ and $C(P)$ be the specialisation and/or generalisation constraints derivable for $H$ and $P$, respectively.
If neither of 
(i) \emph{$P$ is a specialisation of $H$, $H$ is incomplete and $P$ is not inconsistent,} or
(ii) \emph{$P$ is a generalisation of $H$, $H$ is inconsistent and $P$ is not incomplete},
apply, then $\mathcal{H}_{C(H)\cup{}C(P)} \subset \mathcal{H}_{C(H)}$, i.e.~constraints derived for $P$ prune programs not pruned by constraints derived for $H$.
\end{theorem}
\begin{proof}
By case distinction on how $P$ and $H$ are related by subsumption. Note that because $P \neq H$, either $P$ and $H$ are not related by subsumption, or $P$ subsumes $H$, or $H$ subsumes $P$.

Suppose $H$ subsumes $P$, i.e.~$P$ is a specialisation of $H$. 
If $H$ is incomplete, then all of $H$'s specialisations can be pruned, which includes $P$ and its specialisations.
Hence if $P$ is only incomplete then no additional pruning can be achieved, which is exception (i).
If $P$ is (additionally) inconsistent, then $P$'s generalisations can be pruned.
In addition to $H$ being among $P$'s generalisations, there are also programs incomparable with $H$ among $P$'s generalisations, so more pruning can be achieved.

Now suppose $P$ subsumes $H$, i.e.~$P$ is a generalisation of $H$. 
If $H$ is inconsistent, then all of $H$'s generalisations can be pruned, which includes $P$ and its generalisations.
Hence if $P$ is only inconsistent then no additional pruning can be achieved, which is exception (ii).
If $P$ is (additionally) incomplete, then $P$'s specialisations can be pruned.
In addition to $H$ being among $P$'s specialisations, there are also programs incomparable with $H$ among $P$'s specialisations, so more pruning can be achieved.

In the remaining case, where $H$ and $P$ are not related by subsumption, it is immediate that the specialisation/generalisation constraints derived for $P$ prune a distinct part of the hypothesis space, e.g.~$H$'s constraints do not prune $P$.
\end{proof}

\section{Failure explanation algorithm}
\label{sec:algo}
\begin{figure}[]
\begin{minipage}{0.50\linewidth}
\begin{myalgorithm}
def failing_subprogs@$^-$@(@$B$@, @$H$@, @$e^-$@):
    @$T$@ = SLD-tree of @$B \cup H \cup \{ \neg e^- \}$@
    subprogs = {}
    for every successful branch @$\lambda$@ of @$T$@:
        @$H'$@ = sub-program of $H$ identified by
              @$H$@'s clauses that occur in @$\lambda$@
        subprogs = subprogs @$\cup$@ @$\{ H' \}$@
    return subprogs
\end{myalgorithm}
\vspace{1.12em}
\end{minipage}
\hspace{5ex}
\begin{minipage}{0.45\linewidth}
\begin{myalgorithm}
def failing_subprogs@$^+$@(@$B$@, @$H$@, @$e^+$@):
    @$T$@ = SLD-tree of @$B \cup H \cup \{ \neg e^+ \}$@
    subprogs = {}
    for every failing branch @$\lambda$@ of @$T$@:
        @$H'$@ = sub-program of @$H$@ identified by
              @$H$@'s literals that occur in @$\lambda$@
        if SLD-res. fails to prove @${B \cup H' \models e^+}$@:
            subprogs = subprogs @$\cup$@ @$\{ H' \}$@
    return subprogs
\end{myalgorithm}
\end{minipage}
\caption{
Identify failing sub-programs from branches in SLD-trees
}
\label{alg:subprograms}
\end{figure}
We now present a method for identifying failing sub-programs.
The approach is based on the observation that branches of an SLD-tree correspond to sub-programs.
Our algorithm identifies clauses responsible for entailing a negative example.
It is when a program fails to prove entailment that our approach distinguishes itself.
Namely, we also identify literals \emph{within} clauses which cause a positive example to not be entailed.
As the presented method relies on SLD-resolution, from this point on we assume left-to-right evaluation of literals within clauses.


\subsection{SLD-trees}

In algorithmic debugging, missing and incorrect answers help characterise which parts of a \emph{debugging tree} are wrong \cite{ADsurvey}.
Debugging trees can be seen as generalising SLD-trees, with the latter representing the search for a refutation \cite{ilp:book}.
We address the failing sub-programs problem by analysing SLD-trees, only identifying a subset of them.
A \emph{branch} in a SLD-tree is a path from the root \emph{goal} to a leaf.
Each goal on a branch has a \emph{selected atom}, on which resolution is performed to derive child goals.
A branch that ends in an empty leaf is called \emph{successful}, as such a path represents a refutation.
Otherwise a branch is \emph{failing}.
Note that selected atoms on a branch identify a subset of the literals of a program.

\subsection{Identifying sub-programs}

Let $B$ be a definite program, $H$ be a hypothesis, and $e$ be a atom\footnote{While in our application to synthesis we only use ground atoms $e$, the failure explanation algorithm presented in this section also works when $e$ is non-ground.}.
The SLD-tree $T$ for $B \cup H \cup \{ \neg e \}$, with $\neg e$ as the root, proves $B \cup H \models e$ iff $T$ contains a successful branch.
Given a branch $\lambda$ of $T$, we define the $\lambda$-sub-program of $H$.
A literal $L$ of $H$ occurs in \emph{$\lambda$-sub-program} $H'$ if and only if $L$ occurs as a selected atom\footnote{Note that resolution might have unified arguments of $L$ to produce the selected atom.} in $\lambda$ or $L$ was used to produce a resolvent that occurs in $\lambda$.
The former case is for literals in the body of clauses and the latter for head literals.
Now consider the SLD-tree $T'$ for $B\cup H' \cup \{ \neg e \}$ with $\neg e$ as root.
As all literals necessary for $\lambda$ occur in $B\cup H'$,
the branch $\lambda$ must occur in $T'$ as well.

Suppose $e^-$ is an incorrect answer for hypothesis $H$.
Then the SLD-tree for $B \cup H \cup \{\neg e^-\}$ has a successful branch $\lambda$.
The literals of $H$ necessary for this branch are also present in $\lambda$-sub-program $H'$,
hence $e^-$ is also an incorrect answer of $H'$.
Now suppose $e^+$ is a missing answer of $H$.
Let $T$ be the SLD-tree for $B \cup H \cup \{\neg e^+\}$ and $\lambda'$ be any failing branch of $T$.
The literals of $H$ in $\lambda'$ are also present in $\lambda'$-sub-program $H''$.
While $\lambda'$ must be a failing branch present in the SLD-tree of $B \cup H'' \cup \{\neg e^+\}$, this is, in general, insufficient for concluding that this SLD-tree has no successful branch.
Hence whether $e^+$ is indeed a missing answer of $H''$ needs to be verified.

Figure \ref{alg:subprograms} shows the corresponding procedures for deriving failing sub-programs, in the case of a negative example and a positive example, respectively.
Note that hypothesis $H$ can refer to library $B$ but $B$ is not allowed to refer to $H$.
Hence whilst resolving a selected literal of $H$ defined by $B$ with clauses of $B$ we cannot encounter literals of $H$.
Therefore, for failure explanation purposes, we need not inspect the part of the SLD-tree for $B \cup H \cup \{ \neg e \}$ that deals with determining whether a literal defined by $B$ holds or not.
This is equivalent to viewing $B$ as a (possibly infinite) set of facts, i.e.~resolving a selected literal defined by $B$ always returns directly.
This is how we will treat resolving literals of $B$ from this point on.

The following example illustrates identifying sub-programs from the SLD-trees of a recursive program.

\begin{example}
\label{ex:sld-tree-example}
Let $H$ be the following recursive \tw{droplast/2} hypothesis, where the name \tw{droplast} has been shortened to \tw{dl}:
\begin{center}
$\begin{array}{l}
c_1: \tw{dl(A,B):- tail(A,B),empty(B).}\\
c_2: \tw{dl(A,B):- tail(A,C),dl(C,B).}\\
\end{array}$
\end{center}

Suppose $B$ includes the usual definitions for \tw{tail/2} and \tw{empty/1}.
Testing whether $B \cup H \models \tw{dl([1,2],[1])}$ holds is done by SLD-resolution.
The SLD-tree for $B \cup H \cup \{\neg\tw{dl([1,2],[1])}\}$ is:

\newcommand{\la}{$\leftarrow$}
{\small
\begin{center}
\begin{tikzpicture}[scale=0.9, every node/.style={scale=0.9}]
\node at (0,0) (root) {\la\tw{\underline{dl([1,2],[1])}}};
\node[anchor=east] at (-0.2,-1) (l) {\textbf{1:}\la\tw{\underline{tail([1,2],[1])},empty([1])}};
\node[anchor=west] at (0.2,-1) (r) {\la\tw{\underline{tail([1,2],C)},dl(C,[1])}};
\node[anchor=west] at (0.2,-2) (rd) {\la\tw{\underline{dl([2],[1])}}};
\node[anchor=east] at (-0.2,-3) (rdl) {\textbf{2:}\la\tw{\underline{tail([2],[1])},empty([1])}};
\node[anchor=west] at (0.2,-3) (rdr) {\la\tw{\underline{tail([2],C)},dl(C,[1])}};
\node[anchor=west] at (0.2,-4) (rdrd) {\la\tw{\underline{dl([],[1])}}};
\node[anchor=east] at (-0.2,-5) (rdrdl) {\textbf{3:}\la\tw{\underline{tail([],[1])},empty([1])}};
\node[anchor=west] at (-0.068,-5) (rdrdr) {\textbf{4:}\la\tw{\underline{tail([],C)},dl(C,[1])}};

\draw (root) -> node[above left] {$c_1$} (l)
      (root) -> node[above right] {$c_2$} (r);
\draw (rd |- r.south) -> (rd);
\draw (rd) -> node[above left=0.01] {$c_1$} (rdl)
      (rd.south) -> node[right] {$c_2$} (rd |- rdr.north);
\draw (rd |- rdr.south) -> (rd |- rdrd.north);
\draw (rdrd) -> node[above left=0.01] {$c_1$} (rdrdl)
      (rd |- rdrd.south) -> node[right] {$c_2$} (rd |- rdrdr.north);

\end{tikzpicture}
\end{center}
}

Each node is a goal and has its selected literal underlined.
The SLD-tree has four branches, each of them failing.
The branch marked `\textbf{1:}' identifies the sub-program $P_1$ = \{ \tw{dl(A,B):- tail(A,B)}. \} as only clause $c_1$ is used and only its head and first body literal are evaluated.
The branches marked `\textbf{2:}' and `\textbf{3:}' identify the sub-program $P_2$ = \{ \tw{dl(A,B):- tail(A,B)}. ~~ \tw{dl(A,B):- tail(A,C),dl(C,B)}. \} as both clauses are used though the second literal of $c_1$ is never selected while all of the literals of $c_2$ are.
The branch marked `\textbf{4:}' never uses clause $c_1$ and hence identifies sub-program $P_3 = \{ c_2 \}$.
Retesting \tw{dl([1,2],[1])} on these sub-programs confirms that they fail.

Now consider testing for $B \cup H \models \tw{dl([1,2],[])}$.
The SLD-tree for $B \cup H \cup \{\neg\tw{dl([1,2],[])}\}$ has failing branches but also a successful one:
\la\tw{dl([1,2],[])} $\overset{c_2}{\textnormal {---}}$ 
\la\tw{tail([1,2],C),dl(C,[])} --- 
\la\tw{dl([2],[])} $\overset{c_1}{\textnormal {---}}$ 
\la\tw{tail([2],[]),empty([])} --- 
\la\tw{empty([])} --- 
$\square$.
As this branch used all clauses, it identifies $H$ itself as responsible.
On the other hand, the SLD-tree for $B \cup H \models \tw{dl([1],[])}$ has a successful branch only using $c_1$:
\la\tw{dl([1],[])} $\overset{c_1}{\textnormal {---}}$ 
\la\tw{tail([1],[]),empty([])} --- 
\la\tw{empty([])} --- 
$\square$.
Hence it identifies $P_4 = \{ c_1 \}$ as the responsible sub-program.
\end{example}


\section{Implementation}
\label{sec:implementation}
\newcommand{\mitr}[0]{\texttt{mi}$_{\textit{tr}}$}

Before introducing our ILP system, \name{}, we discuss our implementation of the failure explanation algorithm.

\subsection{Meta-Interpreter for failure explanation}

We implement our failure explanation algorithm by a meta-interpreter, \mitr{}, where this meta-interpreter is best understood as instrumenting the program such that executing it keeps track of which parts of the program actually got executed.

Given a background knowledge program $B$ and an atom $G$,
\mitr{} keeps track of which literals of a definite program $P$ have been encountered along each branch of the SLD-tree of $B \cup P \cup \{\neg{}G\}$.
For each literal of the hypothesis $P$ being evaluated we keep track of one bit of information: whether this literal\footnote{Note that the meta-interpreter only keeps track of seen literals of the hypothesis, not of any literals occurring in the background knowledge} has been seen along the current branch or not.
\mitr{} maintains a bitset, which we refer to as a \emph{trace}, containing a unique bit for each literal of the hypothesis.

The meta-interpreter assumes a program transformation $X(\cdot)$ has been applied to the program (where, for notational convenience, clauses are represented by disjunctions):
\[
\begin{array}{rcl}
\vspace{1ex}
X(P) &=& \{ X(C,C_\textit{idx}) ~|~ C \in P \} \\
     &=& \{ \bigvee
\left\{\begin{array}{l}
\neg{}X(A,C_\textit{idx},L_\textit{idx})~\textit{if}~L = \neg{}A\\
X(L,C_\textit{idx},L_\textit{idx})\hspace{0.9em}\textit{otherwise} 
\end{array}
\right.
~|~ L \in C \land C \in P \} \\
\end{array}
\]

Before defining $X(\cdot,\cdot,\cdot)$, we specify how bitsets are derived.
$C_\textit{idx}$ and $L_\textit{idx}$ correspond to the index of clause $C$ within $P$ and the index of $L$ within $C$, respectively.
The function \textit{bitset}$(\cdot,\cdot)$ converts a clause index and literal index within that clause to a bitset with a unique bit set for these inputs.
$X(L,C_\textit{idx},L_\textit{idx})$ \texttt{:= mi}$(L,\textit{bitset}(C_\textit{idx},L_\textit{idx}))$, if the predicate of $L$ is defined by $P$.
Otherwise $X(L,C_\textit{idx},L_\textit{idx})$ \texttt{:= call}$(L,\textit{bitset}(C_\textit{idx},L_\textit{idx}))$, i.e.~in the case the predicate of $L$ is defined by the background knowledge.

%
Figure \ref{alg:meta-interpreter} lists the code for meta-interpreter \mitr{}.
Given an atom $G$ and program $X(P)$, we can evaluate $G$ as a goal using the meta-interpreter by invoking \texttt{\mitr{}(mi($G$,0),0,Trace)}, where \texttt{0} denotes the empty bitset.
When this call succeeds, \texttt{Trace} will have become unified with a bitset identifying all literals that occurred on the first successful branch in the SLD-tree of $B \cup P \cup \{\neg{}G\}$.
If evaluation of \texttt{\mitr(mi($G$,0),0,Trace)} fails then there is no successful branch in the SLD-tree of $B \cup P \cup \{\neg{}G\}$.
In this case \mitr{} will have asserted traces for each unsuccessful branch, via a non-logical predicate \texttt{assert\_failed\_trace}\footnote{\textit{Asserting a trace} can be done in constant time, e.g.~by putting the trace in a hashmap or prepending the trace to the front of a list of failed traces.}.
Upon \texttt{\mitr{}(mi($G$,0),0,Trace)} having failed, all these asserted traces can be inspected to obtain the corresponding sub-programs.

\begin{figure}[]
\begin{minipage}{\linewidth}
\begin{myalgorithm}
mi@$_{\textit{tr}}$@(true,Trace,Trace).
mi@$_{\textit{tr}}$@((HeadOfBody,TailOfBody),Tr@$_{\textit{in}}$@,Tr@$_{\textit{out}}$@) :-
    mi@$_{\textit{tr}}$@(HeadOfBody,Tr@$_{\textit{in}}$@,Tr@$_{\textit{mid}}$@),
    mi@$_{\textit{tr}}$@(TailOfBody,Tr@$_{\textit{mid}}$@,Tr@$_{\textit{out}}$@).
mi@$_{\textit{tr}}$@(mi(G,I),Tr@$_{\textit{in}}$@,Tr@$_{\textit{out}}$@) :-
    clause(mi(G,J),Body),
    Tr@$_{\textit{head}}$@ is Tr@$_{\textit{in}}$@ @$\vee$@ I @$\vee$@ J,
    mi@$_{\textit{tr}}$@(Body,Tr@$_{\textit{head}}$@,Tr@$_{\textit{out}}$@).
mi@$_{\textit{tr}}$@(call(G,I),Tr@$_{\textit{in}}$@,Tr@$_{\textit{out}}$@) :-
    Tr@$_{\textit{out}}$@ is Tr@$_{\textit{in}}$@ @$\vee$@ I,
    (call(G) *-> true ; assert_failed_trace(Tr@$_{\textit{out}}$@),fail).
\end{myalgorithm}
\end{minipage}
\caption{
Meta-interpreter \texttt{mi}$_\textit{tr}$. \texttt{mi}$_\textit{tr}$ keeps track of a trace of literal indices encountered along each SLD-branch. 
The $\vee$ operator takes two bitsets and produces their union (like taking the logical \textit{or} of two integers).
\texttt{call(G)} just interpreters (complex) term \texttt{G} as an atom and evaluates it.
The semantics of \texttt{G *-> Then ; Else} are that if \texttt{G} ever succeeds the entire construct acts as if it were \texttt{G,Then}, otherwise it acts as if it just were \texttt{Else}.
\texttt{clause(Head,Body)} unifies with any definite clause the Prolog interpreter knows about.
\texttt{Body} is a cons-list of atoms which terminates in \texttt{true}.
}
\label{alg:meta-interpreter}
\end{figure}

Note that \mitr{} only does a constant number of additional (bitset unioning / logical \textit{or}) operations at every node of the SLD-tree of $B \cup P \cup \{\neg G \}$ involving literals of $H$ (that is, resolving literals defined $B$ is relegated to the normal interpreter).
Hence the SLD-tree of $B \cup X(P) \cup \{ \neg$\texttt{\mitr(mi($G$,0),0,Trace)}$ \}$ is only a constant factor bigger than the original.
It follows that the overhead \mitr{} incurs from identifying sub-programs is directly proportional to the size of the SLD-tree generated during normal execution, i.e.~the algorithm for identifying sub-programs has linear complexity (and leaves the part of the SLD-tree which is resolving literals of $B$ with clauses of $B$ untouched, incurring no overhead).
This approach does not address non-termination issues of (recursive) programs, i.e.~if executing the original program led to an infinite branch in the SLD-tree then executing the meta-interpreter instead will also yield an infinite branch. 
For sub-programs identified on missing answers, we still need to re-evaluate the sub-programs.
If $P = \{ C_1, \ldots, C_n \}$, then there are $\prod_{1 \leq i \leq n} \#literals(C_i)$ distinct sub-programs of $P$, i.e.~the possible combinations of prefixes of $P$'s clauses, that could be identified for retesting.

\subsection{\name{}}

We now introduce \name{}, an ILP system based on \popper{} \cite{popper}, which supports failure explanation.
\name{} tackles the LFF problem (Definition \ref{def:probin}) using a \emph{generate}, \emph{test}, and \emph{constrain} loop.
\name{} maintains a logical formula (expressed as an answer set program) whose models correspond to the viable hypotheses, i.e.~each model represents a unique Prolog program.

The generate stage is identical to that of \popper{} and searches for a model of the formula which it converts to a program.
In the test stage, a thus generated hypothesis $H$ is tested on positive and negative examples.
\name{} incorporates Algorithm \ref{alg:subprograms}, running it for each tested example.
Meta-interpreter \mitr{} is used to determine clauses and literals that occur along branches responsible for a failure.
From this information \name{} reconstructs the corresponding sub-programs.
If sub-program $H'$ is derived from a branch for a missing answer, $H'$ gets retested, this time using standard SLD-resolution.
The test stage tells the constrain stage the number of missing and incorrect answers of a (sub-)program.
This determines whether its specialisations%
\footnote{\popper{} and \name{} generate \emph{elimination constraints} when a hypothesis entails none of the positive examples \cite{popper}.}
and/or generalisations should be pruned.
For each failed hypothesis and each of its failing sub-programs, new hypothesis constraints are added to the formula, eliminating models, thereby pruning the hypothesis space.
\rolf{AAAI R1: "The paper should discuss what happens when different examples present a conflict between possible search space pruning. For example, if an example points out the need to shorten the clause, others would benefit from adding a larger clause."}
As in general failing sub-programs need not be specialisations/generalisations of $H$,
pruning for sub-programs is in addition to the pruning which the constrain stage already does for $H$ in \popper{}.
Finally, \name{} loops back to the generate stage.
\rolf{AAAI R1 wants it spelled out what happens if no complete and consistent program in hypothesis space.}

Smaller programs prune more effectively, which is partly why \popper{} and \name{} search for hypotheses by increasing size\footnote{The other reason is to find \emph{optimal} solutions, i.e.~those with the minimal number of literals.} (in terms of number of literals).
Yet there are many small programs that \popper{} does not consider well-formed that lead to significant pruning.
Consider the sub-program $H_1' = \{~\emph{droplast(A,B) $\leftarrow$ empty(A)}~\}$ from Example \ref{ex:explain_failures}.
\popper{} does not generate $H_1'$ as it does not consider it a well-formed hypothesis (as the head variable $\emph{B}$ does not occur in the body).
Yet precisely because this sub-program has so few body literals is why it is so effective at pruning specialisations.

The following example demonstrates the loop used by \name{} and \popper{}, and how failure explanation can lead to fewer loop iterations.

\begin{example}
\label{ex:loop}

\begin{figure*}[t]
\centering
\begingroup
\newcommand{\hypo}[2]{{
\tw{#1} = \left\{
\begin{array}{l}
#2
\end{array}
\right\}
}\\
}
\[
\mathcal{H}_1 = \left\{
\begin{array}{l}

\hypo{h$_1$}{
\tw{droplast(A,B):- empty(A),tail(A,B).}\\
}

\hypo{h$_2$}{
\tw{droplast(A,B):- empty(A),cons(C,D,A),tail(D,B).}\\
}

\hypo{h$_3$}{
\tw{droplast(A,B):- tail(A,C),tail(C,B).}\\
\tw{droplast(A,B):- tail(A,B).}\\
}

\hypo{h$_4$}{
\tw{droplast(A,B):- empty(A),tail(A,B),head(A,C),head(B,C).}\\
}

\hypo{h$_5$}{
\tw{droplast(A,B):- tail(A,C),tail(C,B).}\\
\tw{droplast(A,B):- tail(A,B),tail(B,A).}\\
}

\hypo{h$_6$}{
\tw{droplast(A,B):- tail(A,B),empty(B).}\\
\tw{droplast(A,B):- cons(C,D,A),droplast(D,E),cons(C,E,B).}\\
}

\hypo{h$_7$}{
\tw{droplast(A,B):- tail(A,C),tail(C,B).}\\
\tw{droplast(A,B):- tail(A,B).}\\
\tw{droplast(A,B):- tail(A,C),droplast(C,B).}\\
}

%
%
%
\end{array}
\right\}
\]
\endgroup
\caption{LFF hypothesis space considered in Example \ref{ex:loop}.}
\label{fig:hypospace}
\end{figure*}

We illustrate \name{}, and how it differs from \nameold{}, by running its loop on LFF input $(E^+, E^-, \mathcal{H}, B, C)$ from Example \ref{ex:lff_input}.
For demonstration purposes we use the simplified hypothesis space $\mathcal{H}_1 \subseteq \mathcal{H}_{C}$ of Figure \ref{fig:hypospace}.
Our positive examples are
$e_1^+ = \emph{droplast}([1,2,3],[1,2])$
and
$e_2^+ = \emph{droplast}([1,2],[1])$,
and our negative example is
$e_1^- = \emph{droplast}([1,2],[])$.

First we induce a program by a generate-test-and-constrain loop \emph{without} failure explanation.
This first sequence is representative of \popper{}'s execution:

\begin{enumerate}
\item
\popper{} starts by generating \tw{h$_1$}.
$B \cup \tw{h$_1$}$ fails to entail $e_1^+$ and $e_2^+$ and correctly does not entail $e_1^-$.
Hence only specialisations of \tw{h$_1$} are pruned, namely \tw{h$_4$}.
\item
\popper{} subsequently generates \tw{h$_2$}.
$B \cup \tw{h$_2$}$ fails to entail $e_1^+$ and $e_2^+$ and is correct on $e_1^-$.
Hence specialisations of \tw{h$_2$} are pruned, of which there are none in $\mathcal{H}_1$.
\item
\popper{} next generates \tw{h$_3$}.
$B \cup \tw{h$_3$}$ does not entail the positive examples, but does entail negative example $e_1^-$.
Hence specialisations and generalisations of \tw{h$_3$} are pruned, meaning only generalisation \tw{h$_7$}.
\item
\popper{} generates \tw{h$_5$}.
$B \cup \tw{h$_5$}$ is correct on none of the examples.
Hence specialisations and generalisations of \tw{h$_5$} are pruned, of which there are none in $\mathcal{H}_1$.
\item
\popper{} generates \tw{h$_6$}.
$B \cup \tw{h$_6$}$ is correct on all the examples and hence \tw{h$_6$} is returned.
\end{enumerate}

%
%
%
%

\noindent
Now we consider learning by a generate-test-and-constrain loop \emph{with} failure explanation.
The following execution sequence is representative of \name{}:

\begin{enumerate}
\item
\name{} starts by generating \tw{h$_1$}.
$B \cup \tw{h$_1$}$ fails to entail $e_1^+$ and $e_2^+$ and correctly does not entail $e_1^-$.
Failure explanation identifies sub-program $\tw{h$_1'$} = \{ \tw{droplast(A,B):- empty(A).} \}$.
\tw{h$_1'$} fails in the same way as \tw{h$_1$}.
Hence specialisations of both \tw{h$_1$} and \tw{h$_1'$} get pruned, namely \tw{h$_2$} and \tw{h$_4$}.
\item
\name{} subsequently generates \tw{h$_3$}.
$B \cup \tw{h$_3$}$ does not entail the positive examples, but does entail negative example $e_1^-$.
Failure explanation identifies sub-program $\tw{h$_3'$} = \{ \tw{droplast(A,B):- tail(A,C),tail(C,B).} \}$.
$B \cup \tw{h$_3'$}$ fails in the same way as \tw{h$_3$}.
Hence specialisations and generalisations of \tw{h$_3$} and \tw{h$_3'$} get pruned, meaning \tw{h$_5$} and \tw{h$_7$}.
\item
\name{} next generates \tw{h$_6$}.
$B \cup \tw{h$_6$}$ is correct on all the examples and hence \tw{h$_6$} is returned.
\end{enumerate}

\noindent
The difference in these two execution sequences is illustrative of how failure explanation, by way of sub-programs, can help prune away significant parts of the hypothesis space.
\end{example}

\section{Experiments}
\label{sec:experiments}

We claim that failure explanation can improve learning performance.
Our experiments therefore aim to answer the questions:

\begin{itemize}
\item[\textbf{Q1}] Can failure explanation prune more programs?
\item[\textbf{Q2}] Can failure explanation reduce learning times?
\end{itemize}

\noindent
Note that an affirmative answer to \textbf{Q1} does not imply that \textbf{Q2} is the case, as potentially the overhead of failure explanation exceeds the benefits of the pruning it achieves.
\rolf{In order to gain an understanding of the overhead involved in failure explanation, we also aim for an answer to the following question:}
\rolf{
Q3: How does our failure explanation algorithm scale with the size of the SLD-tree?
}
\rolf{To help answer Q3, we run a program for which we know that the SLD-tree scales with the size of an argument of the goal.}
\ac{do compare against normal prolog eval}
\ac{eval how failure explanation scales with number of examples}
\rolf{eval the cost of retesting sub-programs}


To answer \textbf{Q1} and \textbf{Q2}, we compare \name{} against \popper{}.
The addition of failure explanation is the only difference between the systems.
In each of the experiments, the settings for \name{} and \popper{} are identical.
Though control over a system's failure explanation capabilities is required to help answer \textbf{Q1} and \textbf{Q2}, we nevertheless include a comparison against state-of-the-art ILP system Metagol \cite{metagol} 
and the classical ILP system Aleph \cite{aleph}.

We run the experiments on a 10-core server (at 2.2GHz) with 30 gigabytes of memory (note that all the systems only run on a single CPU).
When testing individual examples, we use an evaluation timeout of 2 milliseconds.

\subsection{Experiment 1: robot route planning}
\label{sec:expr-robots}

We first evaluate the potential performance improvement of failure explanation as a function of target program size.
We select a contrived setting where failure explanation ought to be very effective: 
a basic route planning problem.
A robot resides in a grid world and can move in four directions.
The robot starts in the lower left corner and needs to move to a position to its right.
Unbeknownst to the robot, it has been restricted to a corridor (dimensions $14 \times 1$).
In this experiment, failure explanation should determine that any strategy that moves up, down, or starts by moving left can never succeed.

\paragraph{Settings.}
An example is an atom $f(s_1,s_2)$, with start ($s_1$) and end ($s_2$) states.
A state is a pair of discrete coordinates $(x,y)$.
We provide four dyadic relations as BK: $\emph{move\_right}$, $\emph{move\_left}$, $\emph{move\_up}$, and $\emph{move\_down}$, which change the state, e.g.~$\emph{move\_right((2,2),(3,2))}$.
We ensure that our hypotheses are forward-chained \cite{hexmil}, meaning body literals modify the state one after another.
We supply \metagol{} with the following metarules: $P(A,B)\leftarrow Q(A,B)$ and $P(A,B)\leftarrow Q(A,C),R(C,B)$ and $P(A,B) \leftarrow Q(B,A)$.

\paragraph{Systems.}

\ac{I now can see why a reviewer would find metagol no reuse system to be very confusing.
I would drop it and run default metagol (i.e. with PI) and say that it shows the benefits of PI for reuse.}
In comparing systems, we try to ensure that hypothesis spaces are as similar as possible.
For \name{}, \popper{} and \alephilp{} we allow one clause with up to 13 body literals and 14 variables.
\metagol{} is the only system that uses predicate invention, i.e.~learns clauses with invented predicate symbols.
As reusing invented predicates leads to exponentially shorter programs for this problem, we use both \metagol{} and
a version of \metagol{} where reuse of invented predicates is disabled: \metagoln{}.

\paragraph{Method.}
The start state is $(0,0)$ and the end state is $(n,0)$, for $n$ in $1,2,3,\ldots,13$.
Each trial has only one (positive) example: $f((0,0),(n,0))$.
We measure learning times and, for \nameold{} and \name{}, the number of generated programs.
We enforce a timeout of 60 seconds per task.
We repeat each experiment 10 times and plot the mean and standard error.

\paragraph{Results.}
Figure \ref{fig:robots-time} shows that \name{} substantially outperforms \nameold{} in terms of learning time.
The reason for the improved learning time is that \name{} generates far fewer programs, see Figure \ref{fig:robots-generated}.
For example, upon \name{} generating one program that starts by moving left, failure explanation determines any program whose first move is to the left is going to fail and hence all these programs get pruned.
%

Figure \ref{fig:robots-time} also shows that \name{} outperforms \metagoln{}.
Because \metagoln{} is example-driven it is effective in pruning programs that try to move out of the corridor.
Yet, as explained in Section 2, at bigger program sizes its reconsidering of already seen programs is very costly.

\alephilp{} and normal \metagol{} always find the solution, even at size 13, witin 1.5 seconds.
For \metagol{}, this is due to reusing invented predicates.
For example, the size 12 solution that \metagol{} finds has only eight body literals, versus the 12 that \name{} needs.
For \alephilp{}, the bottom-clause construction is very effective in only considering moves that are actually allowed.
However, the performance of these systems does not have bearing on whether failure explanation is effective or not.

The results from this simple experiment strongly suggest that the answer to questions \textbf{Q1} and \textbf{Q2} is yes.

\begin{figure}[t]
\small
\centering
\begin{tikzpicture}
\begin{customlegend}[legend columns=6,legend style={nodes={scale=0.5, transform shape},align=left,column sep=0ex},
        legend entries={\name{}, \popper{}, \metagoln, \metagol{}, \alephilp{}}]
        \addlegendimage{mark=*,blue}
        \addlegendimage{mark=+,red}
        \addlegendimage{mark=diamond*,orange,dashed}
        \addlegendimage{mark=+,black,dashed}
        \addlegendimage{mark=*,green,dashed}
\end{customlegend}
\end{tikzpicture}
\\
\begin{subfigure}[b]{0.45\textwidth}
\centering
\pgfplotsset{every tick label/.append style={font=\Large}}
\begin{tikzpicture}[scale=.48]
   \begin{axis}[
   xlabel=Program size,
   ylabel=Learning time (seconds),
   xmin=1,xmax=13,
   ymin=0,ymax=60,
   ylabel style={xshift=1mm,yshift=-2mm},
   label style={font=\LARGE},
   ]

\addplot[blue,mark=*,mark options={fill=blue},error bars/.cd,y dir=both,y explicit]
table [
x=size,
y=time,
y error plus expr=\thisrow{error},
y error minus expr=\thisrow{error},
] {
size time error
1 0.6581787824630737 0.03123157701647036
2 0.6743480443954468 0.02800466783201579
3 0.7160430669784545 0.01601440811418894
4 0.8338706493377686 0.012702389085220906
5 1.2322434186935425 0.12315583315557234
6 2.2401121616363526 0.45002047089108804
7 5.244803619384766 0.9144187517036124
8 11.522005987167358 0.8312278553096677
9 19.49911026954651 0.533526129339784
10 27.286946177482605 1.2581410199120253
11 36.01977152824402 1.1973330896162449
12 50.19688582420349 1.4709594870616147
13 70 0
};

\addplot[red,mark=+,error bars/.cd,y dir=both,y explicit]
table [
x=size,
y=time,
y error plus expr=\thisrow{error},
y error minus expr=\thisrow{error},
] {
size time error
1 0.6448334217071533 0.02489766612041678
2 0.6700809001922607 0.01913602583441222
3 0.7261670351028442 0.032784853620207305
4 1.1942937612533568 0.017207269296510442
5 4.1347492933273315 1.1318403046906604
6 20.99340660572052 3.8997864891939855
7 60.35367250442505 0.004076856971634039
};


\addplot[orange,mark=diamond*,dashed,error bars/.cd,y dir=both,y explicit]
table [
x=size,
y=time,
y error plus expr=\thisrow{error},
y error minus expr=\thisrow{error},
] {
size time error
1 0.44961984157562257 0.010210368482281278
2 0.45094382762908936 0.010285900196334026
3 0.45924108028411864 0.014959344832520717
4 0.45920183658599856 0.01743997848816055
5 0.4641364336013794 0.0065645444958469905
6 0.6183980226516723 0.013700215262123654
7 5.187635278701782 0.030748701732328177
8 60.273610734939574 0.051584333104189195
};



\addplot[green,dashed,mark=*,error bars/.cd,y dir=both,y explicit]
table [
x=size,
y=time,
y error plus expr=\thisrow{error},
y error minus expr=\thisrow{error},
] {
size time error
1 0.5930155754089356 0.02362037230033172
2 0.6016797542572021 0.0175119123466642
3 0.6023468494415283 0.017466825509079142
4 0.6091533899307251 0.021648410178031454
5 0.5959182500839233 0.013656332511546573
6 0.610544753074646 0.02283568753205451
7 0.6033302068710327 0.014132084664009868
8 0.6218339920043945 0.013757411634706444
9 0.6393617391586304 0.01818612499638782
10 0.6877244234085083 0.02721048833415218
11 0.7813033580780029 0.03176187881486881
12 0.9067548274993896 0.031467901033204314
13 1.1463078737258912 0.05720775361476661
};

\addplot[black,dashed,mark=+,error bars/.cd,y dir=both,y explicit]
table [
x=size,
y=time,
y error plus expr=\thisrow{error},
y error minus expr=\thisrow{error},
] {
size time error
1 0.43108062744140624 0.013273804724063216
2 0.4230064868927002 0.005482609666644143
3 0.450657320022583 0.02528234534824606
4 0.43714404106140137 0.009387506476232118
5 0.437707781791687 0.010978453418915218
6 0.4418792247772217 0.019391691597863973
7 0.44393281936645507 0.012385893951408126
8 0.43986289501190184 0.009807297385518436
9 0.4479323625564575 0.010571755163606257
10 0.4600589036941528 0.017925175337817345
11 0.6202880382537842 0.02490220866019846
12 0.47993299961090086 0.018348902969066452
13 0.6179438591003418 0.01286925538801702
};

\legend{}
\end{axis}
\end{tikzpicture}
\caption{
Learning time.
}
\label{fig:robots-time}
\end{subfigure}
\hspace{0ex}
\begin{subfigure}[b]{0.45\textwidth}
\centering
\pgfplotsset{every tick label/.append style={font=\Large}}
\begin{tikzpicture}[scale=.48]
   \begin{axis}[
   xlabel=Program size,
   ylabel=Generated programs,
   xmin=1,xmax=13,
   ymin=0,ymax=6000,
   ylabel style={xshift=0mm,yshift=3mm},
   label style={font=\LARGE},
   legend style={at={(-0.40,40.97)},style={font=\Large,nodes={right}}}
   ]

\addplot[blue,mark=*,mark options={fill=blue},error bars/.cd,y dir=both,y explicit]
table [
x=size,
y=progs,
y error plus expr=\thisrow{error},
y error minus expr=\thisrow{error},
] {
size progs error
1 4.0 0.0
2 10.0 0.0
3 18.0 0.0
4 26.2 0.39999999999999997
5 46.9 5.539855593785816
6 74.8 12.367699866992243
7 143.3 21.076289996107
8 278.0 20.228692493584454
9 526.2 15.848028268526024
10 973.7 37.826049225368486
11 1806.0 117.04699910719624
12 3524.6 56.72248231521607
13 5265.8 509.45987084362196
};

\addplot[red,mark=+,error bars/.cd,y dir=both,y explicit]
table [
x=size,
y=progs,
y error plus expr=\thisrow{error},
y error minus expr=\thisrow{error},
] {
size progs error
1 4.0 0.0
2 11.0 0.0
3 23.0 0.0
4 89.0 0.0
5 265.9 59.92236644192216
6 817.7 117.51174409394153
7 1720.9 5.593746508378799
8 7000 0
};

\legend{}
\end{axis}
\end{tikzpicture}
\caption{%
 Number of programs.
}
\label{fig:robots-generated}
\end{subfigure}

\caption{Results of robot planning experiment.
The x-axes denote the number of body literals in the solution, i.e.~the number of moves required.
Standard error is plotted but is always negligible for \name{}.
}

\label{fig:robots}
\end{figure}

\subsection{Experiment 2: programming puzzles}
\label{expr:lists}

This experiment evaluates whether failure explanation can improve performance when learning programs for recursive list problems, which  other state-of-the-art ILP systems \cite{ilasp3,dilp,hexmil} struggle to solve.
We show that \name{} can drastically outperform \nameold{}, \metagol{} and \alephilp{} on the same 10 problems used to evaluate \nameold{} \cite{popper}, plus three additional ones: \emph{reverse}, \emph{odd1even2}, \emph{sumlist}.

\paragraph{Settings.}
We provide as BK the monadic relations \emph{empty}, \emph{zero}, \emph{one}, \emph{even}, \emph{odd},
the dyadic relations \emph{element}, \emph{head}, \emph{tail}, \emph{increment}, \emph{decrement}, \emph{geq}, and the triadic relations \emph{cons}, \emph{snoc}, \emph{sum}.
With a single fixed hypothesis space for these problems, \nameold{} exhibits significant variance between learning times across problems (ranging from sub-second times for at least four problems to many minutes on others).
To control for this variance, we select hypothesis space settings on a per problem basis, such that \nameold{} has to do non-trivial search but can still find solutions for each problem within the timeout.
See Appendix \ref{sec:puzzlesettings} for the exact settings.


\paragraph{Systems.}
For \name{} and \nameold{}, we provide simple types and mark arguments of predicates as either input or output.
For Metagol, we use the same metarules used to evaluate it against Popper \cite{popper}, listed in Appendix \ref{app:metarules}.
Because \metagol{} uses metarules and invented predicates, its hypothesis space is similar but not identical to that of \name{} and \nameold{}.
For \alephilp{} we provide mode declarations and determinations which encode the exact same information made available to \name{}.
We use the same \alephilp{} settings used to compare it against \nameold{} \cite{popper}: we set the maximum variable depth and clause length to six and the number of search nodes is limited to 30000.


\paragraph{Method.}
We generate 10 positive and 10 negative examples per problem.
Each example is randomly generated from lists up to length 50, whose integer elements are sampled from 1 to 100.
We test on 100 positive and 100 negative randomly sampled examples, giving a default accuracy of 50\%.
We measure learning time, number of programs generated and predictive accuracy.
We also measure the time spent in the three distinct stages of \nameold{} and \name{}.
We repeat each experiment 20 times and record the mean and standard error.
We enforce a 60 second timeout. 

\paragraph{Results.}
\newcommand\narrowc{@{\hskip2pt}c@{\hskip2pt}}

\begin{table}[ht]
\scriptsize
\centering
\begin{tabular}{
\narrowc
\narrowc|\narrowc|\narrowc
\narrowc|\narrowc|\narrowc
\narrowc|\narrowc
}
\multicolumn{1}{c}{} &
\multicolumn{3}{c}{\textbf{Number of programs}} &
\multicolumn{3}{c}{\textbf{Learning time} (sec)} &
\multicolumn{2}{c}{\textbf{Accuracy}}
\\

\cmidrule(r){2-4}
\cmidrule(r){5-7}
\cmidrule(r){8-9}
\textbf{Problem} &
\nameold{} & \name{} & \textbf{ratio} &
\nameold{} & \name{} & \textbf{ratio} &
\nameold{} & \name{} \\
\cmidrule(lr){1-1}
\cmidrule(r){2-4}
\cmidrule(r){5-7}
\cmidrule(r){8-9}

dropk &
2585 $\pm$ 184 & 121 $\pm$ 57 & \textbf{0.05} &
35 $\pm$ 6 & 4 $\pm$ 2 & \textbf{0.11} & 99 $\pm$ 3 & 99 $\pm$ 3 \\
sumlist &
2619 $\pm$ 23 & 127 $\pm$ 17 & \textbf{0.05} &
47 $\pm$ 3 & 3 $\pm$ 0.7 & \textbf{0.07} & 100 $\pm$ 0 & 100 $\pm$ 0 \\
len &
2826 $\pm$ 19 & 172 $\pm$ 18 & \textbf{0.06} &
50 $\pm$ 3 & 3 $\pm$ 0.4 & \textbf{0.06} & 100 $\pm$ 0 & 100 $\pm$ 0 \\
last &
477 $\pm$ 91 & 63 $\pm$ 25 & \textbf{0.13} &
13 $\pm$ 4 & 2 $\pm$ 0.6 & \textbf{0.15} & 100 $\pm$ 0 & 100 $\pm$ 0 \\
droplast &
1718 $\pm$ 117 & 242 $\pm$ 75 & \textbf{0.14} &
41 $\pm$ 8 & 7 $\pm$ 2 & \textbf{0.18} & 100 $\pm$ 0 & 100 $\pm$ 0 \\
odd1even2 &
1324 $\pm$ 272 & 289 $\pm$ 98 & \textbf{0.22} &
17 $\pm$ 5 & 4 $\pm$ 2 & \textbf{0.26} & 100 $\pm$ 0 & 100 $\pm$ 0 \\
member &
173 $\pm$ 36 & 64 $\pm$ 13 & \textbf{0.37} &
31 $\pm$ 10 & 17 $\pm$ 6 & \textbf{0.54} & 100 $\pm$ 0 & 100 $\pm$ 0 \\
threesame &
136 $\pm$ 44 & 72 $\pm$ 41 & \textbf{0.53} &
10 $\pm$ 6 & 5 $\pm$ 4 & \textbf{0.50} & 100 $\pm$ 0 & 100 $\pm$ 0 \\
finddup &
1167 $\pm$ 82 & 653 $\pm$ 51 & \textbf{0.56} &
10 $\pm$ 1 & 7 $\pm$ 0.6 & \textbf{0.66} & 99 $\pm$ 1 & 99 $\pm$ 1 \\
addhead &
71 $\pm$ 24 & 41 $\pm$ 16 & \textbf{0.57} &
5 $\pm$ 2 & 5 $\pm$ 2 & \textbf{0.98} & 100 $\pm$ 0 & 100 $\pm$ 0 \\
sorted &
861 $\pm$ 221 & 712 $\pm$ 148 & \textbf{0.83} &
32 $\pm$ 12 & 28 $\pm$ 8 & \textbf{0.87} & \textbf{99 $\pm$ 4} & 98 $\pm$ 5 \\
reverse &
1227 $\pm$ 424 & 1025 $\pm$ 435 & \textbf{0.84} &
29 $\pm$ 8 & 28 $\pm$ 10 & \textbf{0.97} & 100 $\pm$ 0 & 100 $\pm$ 0 \\
evens &
786 $\pm$ 7 & 754 $\pm$ 9 & \textbf{0.96} &
14 $\pm$ 0.9 & 16 $\pm$ 0.9 & 1.14 & 100 $\pm$ 0 & 100 $\pm$ 0 \\

\cmidrule(lr){1-1}
\cmidrule(r){2-4}
\cmidrule(r){5-7}
\cmidrule(r){8-9}
\end{tabular}
\caption{
Results for \name{} and \nameold{} for Experiment 2.
Left, the average number of programs generated by each system.
Middle, the (corresponding) average time to find a solution.
Right, the average accuracy of solutions.
The error is standard error.
We round values over one to the nearest integer.
Values under one we round to the most significant digit.
}
\label{tab:puzzles}
\end{table}

\begin{table}[ht]
\scriptsize
\centering
\begin{tabular}{
\narrowc
\narrowc|\narrowc|\narrowc
\narrowc|\narrowc|\narrowc
}
\multicolumn{1}{c}{} &
\multicolumn{3}{c}{\textbf{Number of programs}} &
\multicolumn{3}{c}{\textbf{Total time} (sec)}
\\

\cmidrule(r){2-4}
\cmidrule(r){5-7}
\textbf{Problem} &
\nameold{} & \name{} & \textbf{ratio} &
\nameold{} & \name{} & \textbf{ratio} \\
\cmidrule(lr){1-1}
\cmidrule(r){2-4}
\cmidrule(r){5-7}

addhead* &
42 $\pm$ 0.0 & 25 $\pm$ 0.8 & \textbf{0.58} &
5 $\pm$ 0.1 & 4 $\pm$ 0.2 & \textbf{0.87} \\
reverse* &
770 $\pm$ 2 & 539 $\pm$ 7 & \textbf{0.70} &
20 $\pm$ 0.9 & 17 $\pm$ 0.9 & \textbf{0.83} \\
sorted* &
599 $\pm$ 15 & 477 $\pm$ 9 & \textbf{0.80} &
21 $\pm$ 2 & 18 $\pm$ 1 & \textbf{0.85} \\
\cmidrule(lr){1-1}
\cmidrule(r){2-4}
\cmidrule(r){5-7}
\end{tabular}
\caption{
Selection of programming puzzles for which there was high variance in Table \ref{tab:puzzles}.
Hypotheses spaces for these problems have been pre-pruned of all programs whose size is at least as large as that of the smallest solution.
Total time measures the time, in seconds, required to show there is no solution in these hypothesis spaces.
}
\label{tab:puzzles-nosolution}
\end{table}

\begin{table}[ht]
\scriptsize
\centering
\begin{tabular}{
\narrowc
\narrowc|\narrowc|\narrowc
\narrowc|\narrowc|\narrowc
}
\multicolumn{1}{c}{} &
\multicolumn{3}{c}{\textbf{Learning time} (sec)} &
\multicolumn{3}{c}{\textbf{Accuracy}}
\\

\cmidrule(r){2-4}
\cmidrule(r){5-7}
\textbf{Problem} &
\name{} & \alephilp{} & \metagol{} &
\name{} & \alephilp{} & \metagol{} \\
\cmidrule(lr){1-1}
\cmidrule(r){2-4}
\cmidrule(r){5-7}

dropk &
4 $\pm$ 2 & 7 $\pm$ 18 & N/A & \textbf{99 $\pm$ 2} & 50 $\pm$ 2 & N/A \\

sumlist &
3 $\pm$ 0.7 & 60 $\pm$ 0.0 & N/A & \textbf{100 $\pm$ 0} & 50 $\pm$ 0 & N/A \\

len &
3 $\pm$ 0.4 & 60 $\pm$ 0.1 & 60 $\pm$ 0.1 & \textbf{100 $\pm$ 0} & 50 $\pm$ 0 & 50 $\pm$ 0 \\

last &
2 $\pm$ 0.6 & 1 $\pm$ 0.1 & \textbf{0.7 $\pm$ 0.7} & \textbf{100 $\pm$ 0} & 50 $\pm$ 0 & \textbf{100 $\pm$ 0} \\

droplast &
7 $\pm$ 2 & 60 $\pm$ 0.0 & N/A & \textbf{100 $\pm$ 0} & 50 $\pm$ 0 & N/A \\

odd1even2 &
4 $\pm$ 2 & 56 $\pm$ 9 & 25 $\pm$ 25 & \textbf{100 $\pm$ 0} & 57 $\pm$ 17 & 85 $\pm$ 22 \\

member &
17 $\pm$ 6 & 60 $\pm$ 0.1 & 0.3 $\pm$ 0.0 & \textbf{100 $\pm$ 0} & 50 $\pm$ 0 & 99 $\pm$ 0 \\

threesame &
\textbf{5 $\pm$ 4} & 55 $\pm$ 11 & 5 $\pm$ 12 & \textbf{100 $\pm$ 0} & 60 $\pm$ 20 & \textbf{100 $\pm$ 0} \\

finddup &
7 $\pm$ 0.6 & 1 $\pm$ 0.5 & 2 $\pm$ 2 & 99 $\pm$ 1 & 50 $\pm$ 1 & \textbf{100 $\pm$ 0} \\

sorted &
28 $\pm$ 8 & 0.7 $\pm$ 0.1 & 60 $\pm$ 0.1 & \textbf{98 $\pm$ 5} & 65 $\pm$ 6 & 50 $\pm$ 0 \\

addhead &
5 $\pm$ 2 & 58 $\pm$ 12 & N/A & \textbf{100 $\pm$ 0} & 52 $\pm$ 10 & N/A \\

reverse &
28 $\pm$ 10 & 36 $\pm$ 24 & N/A & \textbf{100 $\pm$ 0} & 50 $\pm$ 0 & N/A \\

evens &
16 $\pm$ 0.9 & 60 $\pm$ 0.1 & 60 $\pm$ 0.1 & \textbf{100 $\pm$ 0} & 50 $\pm$ 0 & 50 $\pm$ 0 \\

\cmidrule(lr){1-1}
\cmidrule(r){2-4}
\cmidrule(r){5-7}
\end{tabular}
\caption{
Results for \name{}, \alephilp{} and \metagol{} for Experiment 2.
On the left the average time to find a solution.
On the right the average accuracy of solutions.
The error is standard error.
We round values over one to the nearest integer.
Values under one we round to the most significant digit.
}
\label{tab:puzzles-systems}
\end{table}

\name{}'s accuracy is at least 98\% on all problems, see Table \ref{tab:puzzles}.
Both \name{} and \nameold{} always terminate before the timeout and score 100\% on the same ten problems.

Table \ref{tab:puzzles} shows the learning times in relation to the number of programs generated.
Crucially, it includes the ratio of the mean of \name{} over the mean of \nameold{}.
On these 13 problems, \name{} always considers fewer hypotheses than \nameold{}.
On seven problems less than 50\% of the original number of programs is considered while only on three problems over 80\% is still needed.

To illustrate why failure explanation is effective, we consider the \emph{dropk} problem.
In a particular run, \nameold{} generates 471 single-clause programs which have \texttt{f(A,B,C):-tail(A,C)} as a sub-program.
On the same examples, \name{} identifies this as a failing sub-program of the first hypothesis it generates and hence immediately prunes all these specialisations.
In total, \nameold{} considers 851 programs with \texttt{f(A,B,C):-tail(A,C)} as a sub-program, whilst \name{} considers just 48.

Failure explanation need not always be effective at pruning.
Consider an arbitrary run of the \textit{evens} problem:
\name{} takes 354 programs before it identifies a sub-program that is not a program it has seen before.
In total \name{} prunes based on just 19 sub-programs.
This can be ascribed to \texttt{evens(A)} being a monadic predicate:
most of the sub-programs that \name{} finds are properly formed \nameold{} programs that \name{} (and \nameold{}) has already seen and learnt constraints from.
On a particular run of \textit{reverse}, \name{} identifies 135 not-before-seen sub-programs.
The first sub-program (of the 5th hypothesis) prunes 112 of \nameold{}'s programs, the second sub-program only 26, the third 15, and from the 5th newly identified sub-program on, which already has four literals, only about three additional programs are pruned versus \nameold{}.
By contrast, the 10th \textit{dropk} sub-program, of size three, still prunes 59 programs relative to \nameold{}.
The effectiveness of failure-explanation-based pruning appears to be strongly dependent on whether many small sub-programs can be identified.

As seen from the ratio columns of Table \ref{tab:puzzles}, the number of generated programs correlates strongly with the learning time (0.96 correlation coefficient).
Only on one problem is \name{} slower than \nameold{}.
Hence outfitting \nameold{} with failure explanation can occasionally affect it negatively, but this result demonstrates that at other times the speed-up can be considerable.

\begin{figure}[t]
\centering
\includegraphics[width=\linewidth]{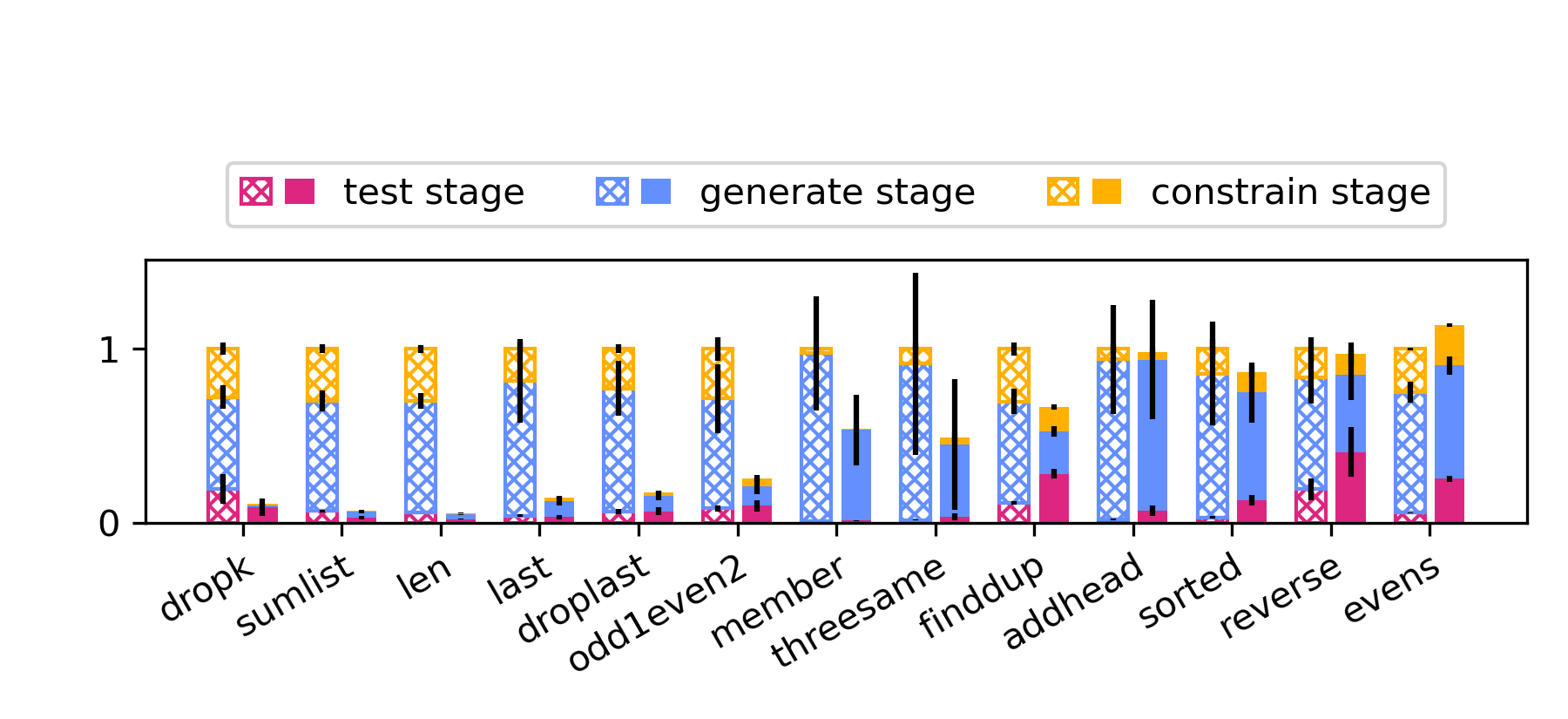}
\caption{
Relative time spent in three stages of \nameold{}, hatched and on the left, and \name{}, on the right.
From bottom to top:
testing, generating hypotheses, and imposing constraints.
Mean times are shown and scaled by the total learning time of \nameold{}.
Bars are standard error.
}
\label{fig:relative_stages}
\end{figure}

Figure \ref{fig:relative_stages} shows the relative time spent in each stage of \name{} and \popper{}.
\rolf{Celine: ``Any insights about the number of constraints generated by the two systems?''
Fair question. This should be expressed in the number of identified sub-programs. Would maybe be interesting to include.
}
%
We can infer the overhead of failure explanation by analysing SLD-trees from this figure.
All problems from \emph{odd1even2} to \emph{evens} have \name{} spend more time on testing than \popper{}.
On \emph{finddup}, \emph{reverse} and \emph{evens}, \name{} incurs considerable testing overhead.
While for \emph{finddup} this effort translates into more effective pruning constraints, for \emph{sorted} and \emph{evens} this is not the case.
Abstracting away from the implementation of failure explanation,
we see that \nameold{} outfitted with zero-overhead failing sub-program identification would have been strictly faster.

There is considerable variance in the number of generated programs and learning times on three problems.
This is in large part due to the solver that is used, Clingo \cite{clingo}, yielding models, i.e.~hypotheses, non-deterministically.
That is, there is no fixed order in which we see hypotheses, so, by chance, \name{} and \nameold{} can come across a solution considerably sooner in one trial than in another.
As a remedy for this variance, we re-run these three problems with their hypothesis spaces restricted to programs that are strictly smaller than solutions.
In this setup, \name{} and \nameold{} always terminate precisely at the point when they have shown that none of these hypotheses can be a solution.
The results, which indeed have less variance, are in Table \ref{tab:puzzles-nosolution}.

Table \ref{tab:puzzles-systems} shows the mean accuracy and learning times of \metagol{} and \alephilp{} versus \name{}.
Accuracy is below 67\% for Aleph on all problems, which can be ascribed to Aleph struggling to learn recursive programs.
\metagol{} cannot find solutions for problems which require arity-three predicates (unless given hand-crafted metarules), which is why `Not Applicable' is listed for five problems.
On another four problems, \metagol{} returns low accuracy hypotheses. 
Only on two problems does \metagol{} outperform \name{}.
In general, \name{} is the more flexible system and outperforms \metagol{} and \alephilp{}.

Overall, these results strongly suggest that the answer to questions \textbf{Q1} and \textbf{Q2} is yes.

\subsection{Experiment 3: IGGP and Michalski trains}

For the next experiment, we evaluate \name{} on problems where solutions are larger, either because they require many clauses or many literals in a clause.
We consider two settings: classification in the form of Michalski train problems \cite{michalski:trains} and inductive general game playing \cite{iggp}.
The problems in these two settings are sufficiently hard that solutions cannot always be found in a reasonable timeframe, hence we rely on \name{}'s anytime capabilities to return the best scoring hypothesis it was able to find upon a timeout.

Michalski train problems concern classifying a train as either eastbound or westbound.
The features available for classifying a train's heading are its cars and their features: if a car is long or short, how many wheels the car has, how many loads and which loads it is carrying, and, finally, whether the car's roof is open, closed or flat.
The target predicate, \texttt{westbound/1}, acts as our classifier and BK predicates allow for inspecting features of the trains to be classified.
We consider the same 4 instances considered by Cropper \cite{dcc-aaai}.
An example of one of the higher quality hypotheses for the \textit{trains4} problem is:

\begin{myalgorithm}
westbound(A):-has_car(A,C),roof_open(C),has_load(C,B),hexagon(B),three_load(B).
westbound(A):-has_car(A,B),has_load(B,D),diamond(D),has_load(B,C),rectangle(C).
\end{myalgorithm}

Inductive General Game Playing concerns learning the rules of games from observations of these games being played.
The goal is to synthesize a set of rules which are consistent with the \emph{traces} generated by a game from the General Game Playing competition \cite{generalgameplaying}.
The four games we consider are: \textit{minimal-decay}, \textit{rock, paper, scissors (rps)}, \textit{buttons} and \textit{coins}.
In each case we learn the predicate \texttt{next}.

\paragraph{Settings \& Systems}
For the trains problems, we provide two dyadic predicates, \texttt{has\_car} and \texttt{has\_load}, and 17 monadic predicates which encode features of cars and loads.
We provide the types of arguments as well as whether they are inputs or outputs to \name{}, \nameold{} and \alephilp{}.
We allow up to four clauses, and within each clause six variables and up to six body literals.
No recursion is allowed.
For \metagol{} we provide the same metarules as in the previous experiment.
For \alephilp{} we limit the search nodes to 30000.

For the IGGP problems we provide the monadic, dyadic and triadic predicates that encode the actions and information available to advance the game to the next state.
For example, for \textit{rps} we look for a definition of \texttt{next\_score/3} given predicates \texttt{true\_score/3}, \texttt{succ/2}, \texttt{does/3}, \texttt{wins/2}, \texttt{beats/2}, \texttt{different/2}.

\paragraph{Method}
We use the same instances of the problems considered by Cropper \cite{dcc-aaai}.
The four \textit{trains} problems represent progressively harder instances, with \textit{trains1} having a one clause six-literal solution and \textit{trains4} needing 26 literals over four clauses for an optimal solution.
Each trains problem has a 1000 examples available, though the distribution between positive and negative varies between tasks.
We follow Cropper in that ``we randomly sample the examples and split them into 80/20 train/test partitions.''
The four games are selected as representative instances of the larger IGGP dataset.

We measure learning time and predictive accuracy.
We repeat each experiment 10 times and record the mean and standard error.
We enforce a 300 second timeout.

\begin{table}[ht]
\scriptsize
\centering
\begin{tabular}{
\narrowc
\narrowc|\narrowc|\narrowc
\narrowc|\narrowc|\narrowc
\narrowc|\narrowc
}
\multicolumn{1}{c}{} &
\multicolumn{3}{c}{\textbf{Number of programs}} &
\multicolumn{3}{c}{\textbf{Learning time} (sec)} &
\multicolumn{2}{c}{\textbf{Accuracy}}
\\

\cmidrule(r){2-4}
\cmidrule(r){5-7}
\cmidrule(r){8-9}
\textbf{Problem} &
\nameold{} & \name{} & \textbf{ratio} &
\nameold{} & \name{} & \textbf{ratio} &
\nameold{} & \name{} \\
\cmidrule(lr){1-1}
\cmidrule(r){2-4}
\cmidrule(r){5-7}
\cmidrule(r){8-9}
rps &
10648 $\pm$ 38 & 250 $\pm$ 13 & \textbf{0.02} &
96 $\pm$ 2 & 25 $\pm$ 1 & \textbf{0.26} & 100 $\pm$ 0 & 100 $\pm$ 0 \\
minimal-decay &
23171 $\pm$ 1538 & 1904 $\pm$ 76 & \textbf{0.08} &
300 $\pm$ 0.0 & 41 $\pm$ 2 & \textbf{0.14} & 94 $\pm$ 0 & \textbf{100 $\pm$ 0} \\
buttons &
8022 $\pm$ 2265 & 1073 $\pm$ 144 & 0.13 &
300 $\pm$ 0.2 & 300 $\pm$ 0.0 & 1.00 & 90 $\pm$ 0 & 90 $\pm$ 0 \\
coins &
9458 $\pm$ 934 & 535 $\pm$ 151 & 0.06 &
300 $\pm$ 0.0 & 300 $\pm$ 0.0 & 1.00 & 88 $\pm$ 3 & 85 $\pm$ 1 \\
\cmidrule(lr){1-1}
\cmidrule(r){2-4}
\cmidrule(r){5-7}
\cmidrule(r){8-9}

trains1 &
28 $\pm$ 0.0 & 20 $\pm$ 0.3 & \textbf{0.72} &
\textbf{1.0 $\pm$ 0.0} & 3 $\pm$ 0.0 & 2.99 & 100 $\pm$ 0 & 100 $\pm$ 0 \\
trains2 &
9410 $\pm$ 6144 & 306 $\pm$ 188 & \textbf{0.03} &
210 $\pm$ 137 & 15 $\pm$ 9 & \textbf{0.07} & 91 $\pm$ 5 & \textbf{98 $\pm$ 2} \\
trains4 &
11223 $\pm$ 377 & 1176 $\pm$ 25 & 0.10 &
300 $\pm$ 0.0 & 300 $\pm$ 0.0 & 1.00 & 78 $\pm$ 2 & \textbf{89 $\pm$ 1} \\
trains3 &
11278 $\pm$ 594 & 1315 $\pm$ 23 & 0.12 &
300 $\pm$ 0.0 & 300 $\pm$ 0.0 & 1.00 & 91 $\pm$ 2 & \textbf{96 $\pm$ 1} \\

\cmidrule(lr){1-1}
\cmidrule(r){2-4}
\cmidrule(r){5-7}
\cmidrule(r){8-9}
\end{tabular}
\caption{
Results for \name{} and \nameold{} for Experiment 3.
Left, the average number of programs generated by each system.
Middle, the (corresponding) average time to find a solution.
Right, the average accuracy of solutions.
The error is standard error.
We round values over one to the nearest integer.
Values under one we round to the most significant digit.
}
\label{tab:iggp-trains}
\end{table}

\begin{table}[ht]
\scriptsize
\centering
\begin{tabular}{
\narrowc
\narrowc|\narrowc|\narrowc
\narrowc|\narrowc|\narrowc
}
\multicolumn{1}{c}{} &
\multicolumn{3}{c}{\textbf{Learning time} (sec)} &
\multicolumn{3}{c}{\textbf{Accuracy}}
\\

\cmidrule(r){2-4}
\cmidrule(r){5-7}
\textbf{Problem} &
\name{} & \alephilp{} & \metagol{} &
\name{} & \alephilp{} & \metagol{} \\
\cmidrule(lr){1-1}
\cmidrule(r){2-4}
\cmidrule(r){5-7}
rps &
25 $\pm$ 1 & \textbf{4 $\pm$ 0.1} & N/A & \textbf{100 $\pm$ 0} & \textbf{100 $\pm$ 0} & N/A \\

minimal-decay &
41 $\pm$ 2 & 4 $\pm$ 0.1 & 300 $\pm$ 0 & \textbf{100 $\pm$ 0} & 94 $\pm$ 0 & 88 $\pm$ 0 \\

buttons &
300 $\pm$ 0 & 137 $\pm$ 4 & 300 $\pm$ 0 & \textbf{90 $\pm$ 0} & 87 $\pm$ 0 & 80 $\pm$ 0 \\

coins &
300 $\pm$ 0 & 300 $\pm$ 0.0 & N/A & \textbf{85 $\pm$ 1} & 82 $\pm$ 0 & N/A \\
\cmidrule(lr){1-1}
\cmidrule(r){2-4}
\cmidrule(r){5-7}
trains1 &
3 $\pm$ 0.0 & \textbf{2 $\pm$ 0.3} & 162 $\pm$ 38 & 100 $\pm$ 0 & 100 $\pm$ 0 & 100 $\pm$ 0 \\

trains2 &
15 $\pm$ 9 & 1 $\pm$ 0.1 & 218 $\pm$ 126 & 98 $\pm$ 2 & \textbf{100 $\pm$ 0} & 85 $\pm$ 6 \\

trains4 &
300 $\pm$ 0 & 215 $\pm$ 4 & 300 $\pm$ 0 & 89 $\pm$ 1 & \textbf{100 $\pm$ 0} & 67 $\pm$ 0 \\

trains3 &
300 $\pm$ 0 & 18 $\pm$ 0.9 & 300 $\pm$ 0 & 96 $\pm$ 1 & \textbf{100 $\pm$ 0} & 20 $\pm$ 0 \\
\cmidrule(lr){1-1}
\cmidrule(r){2-4}
\cmidrule(r){5-7}
\end{tabular}
\caption{
Results for \name{}, \alephilp{} and \metagol{} for Experiment 3.
On the left the average time to find a solution.
On the right the average accuracy of solutions.
The error is standard error.
We round values over one to the nearest integer.
Values under one we round to the most significant digit.
}
\label{tab:iggp-trains-systems}
\end{table}

\paragraph{Results}
Table \ref{tab:iggp-trains} includes the results for \name{} and \nameold{}.
For the IGGP problems, we have that \name{} times out on \textit{coins} and \textit{buttons}, while \nameold{} additionally times out on \textit{minimal-decay}.
On \textit{rps} and \textit{minimal-decay}, \name{} is able to find a solution with 100\% accuracy.
Note how \name{} only required around 250 programs for finding a solution for \textit{rps} while \nameold{} required over 10.000 programs.
For \textit{minimal-decay} \name{} needs to consider almost 2000 programs before coming across a solution while \nameold{} cannot find one within the time limit.
\rolf{There is an issue with soundness on coins. Considering the deadline I am just ignoring it for now.}

In Table \ref{tab:iggp-trains-systems} we see the performance of \metagol{} and \alephilp{} versus \name{} on the IGGP problems.
As \metagol{}'s metarules do not support arity-three predicates, we have that it is unable to find programs for \textit{rps} and \textit{coins}.
On the other two problems, \metagol{} timeouts and hence achieves the default accuracy for these problems.
On \textit{coins}, both \name{} and \alephilp{} achieve the default accuracy.
On \textit{rps}, \alephilp{} does better than \name{} by virtue of its learning time, though \name{} still beats \metagol{}.
On the three other games, \name{} does better than both \alephilp{} and \metagol{}.

Referring back to Table \ref{tab:iggp-trains}, we see that \name{} outperforms \nameold{} on the three more difficult trains problems.
On \textit{trains1} we see clearly the overhead of failure explanation.
Even though \name{} requires less programs than \nameold{}, testing 800 examples incurs 800 times the linear overhead of failure explanation (with regards to SLD-tree size) plus the cost of retesting failing sub-programs, of which there are more when we are dealing with bigger hypotheses.
On the other three problems, the cost of failure explanation is outweighed by the pruning it achieves, with \name{} finding more accurate solutions.
Not shown in Table \ref{tab:iggp-trains}, for the timeouts, \name{} spends a greater proportional of time in the test-stage than \nameold{}, e.g.~about two-thirds of the time on \textit{trains4} versus just one-third of the time, respectively.
This is likely attributable to the cost of retesting many sub-programs on the high number of examples.
\rolf{@AC: note that these experiments were run with \texttt{--test-all} in order to get (meaningful) anytime results.
This is not strictly necessary.
Minimal testing could be made to account for the minimal number of examples that need to be tested to assess whether the current hypothesis is feasible as an improvement over the best seen hypothesis.}

From Table \ref{tab:iggp-trains-systems} we can see that \alephilp{}'s bottom clause construction-based learning procedure is quite effective, outperforming \name{} on all four trains problems.
In turn, \name{} outperforms \metagol{} on all trains problems.

Also for this experiment, the results indicate that the answer to questions \textbf{Q1} and \textbf{Q2} is yes, though with the note that larger hypotheses do appear to impact the effectiveness.

\subsection{Experiment 4: string transformations}
\label{sec:strings}

We now explore whether failure explanation can improve learning performance on real-world string transformation tasks.
We hence restrict ourselves to comparing \name{} versus \nameold{}.
We use a standard dataset \cite{metabias,playgol} formed of 312 tasks, each with 10 input-output pair examples.
For example, task 81 has the following two input-output pairs:

\begin{center}
\begin{tabular}{l|l}
\textbf{Input} & \textbf{Output} \\
\hline
``Alex'',``M'',41,74,170 & M \\
``Carly'',``F'',32,70,155 & F
\end{tabular}
\end{center}

\paragraph{Settings.}
As background knowledge, we give each system the monadic predicates \emph{is\_uppercase},  \emph{is\_empty}, \emph{is\_space}, \emph{is\_letter}, \emph{is\_number} and dyadic predicates \emph{mk\_uppercase}, \emph{mk\_lowercase}, \emph{skip1}, \emph{copyskip1}, \emph{copy1}.
For each monadic predicate we also provide a predicate that is its negation.
%
We allow up to 3 clauses, with each clauses having a maximum of 4 body literals and up to 5 variables.
We extend the test stage with a check whether the generated program is functional or not and prune for any non-functional program.


\paragraph{Method.}
The dataset has 10 positive examples for each problem.
We perform cross validation by selecting 10 distinct subsets of 5 examples for each problem, using the other 5 to test.
We measure learning times and number of programs generated.
We enforce a timeout of 60 seconds per task.
We repeat each experiment 10 times, once for each distinct subset, and record means and standard errors.

\paragraph{Results.}

\begin{figure}[ht]
\centering
\begin{tikzpicture}[scale=0.75]
\begin{axis}[
  xlabel=Ratio of generated programs,
  ylabel=Ratio of learning time,
  ylabel style={yshift=-3mm},
  xlabel style={yshift=1mm},
  xmin=0, xmax=1.3,
  ymin=0, ymax=1.3,
]
\addplot [only marks,mark=*] table {
0.64 0.77
1.00 1.09
0.24 0.33
0.18 0.27
0.26 0.47
0.22 0.24
0.70 0.92
0.65 0.85
0.24 0.42
0.07 0.09
0.72 0.83
0.27 0.55
0.25 0.48
0.45 0.66
0.82 1.10
0.16 0.27
0.25 0.48
0.76 1.07
0.22 0.36
0.55 0.66
0.76 0.94
0.20 0.38
0.20 0.36
0.24 0.45
0.56 0.68
0.18 0.29
0.23 0.43
0.23 0.38
0.22 0.38
0.28 0.54
0.29 0.58
0.23 0.47
0.23 0.46
0.56 0.76
0.24 0.43
0.44 0.78
0.25 0.45
0.27 0.48
0.21 0.42
0.25 0.54
0.24 0.43
0.24 0.41
0.28 0.53
0.60 0.74
0.21 0.46
0.22 0.35
0.24 0.40
0.22 0.35
0.18 0.28
0.21 0.35
0.22 0.38
0.22 0.33
0.25 0.37
0.22 0.34
};
\addplot[dashed, samples=100, smooth,domain=0:1] {x};
\end{axis}
\end{tikzpicture}
\caption{String transformation results.
The ratio of number of programs that \name{} needs versus \nameold{} is plotted against the ratio of learning time needed on that problem.}
\label{fig:strings}
\end{figure}

In 132 problems both \name{} and \nameold{} return programs which have non-zero accuracy on the test set.
On 64 tasks \name{} scores better than \nameold{} versus \nameold{} scoring better on 20 tasks.
For 54 problems at least one of \nameold{} and \name{} finds solutions with over 90\% mean accuracy.
\name{} finds solutions%
\footnote{Note that these problems are very difficult with many of them not having solutions given only our primitive BK and with the learned program restricted to defining a single predicate. Therefore, absolute performance should be ignored. The important result is the relative performance of the two systems.
}
with 100\% accuracy on 37 tasks, 3 more than \nameold{}.

Figure \ref{fig:strings} plots ratios of generated programs and learning times.
Each of the 54 points represents a single problem where either \name{} or \nameold{} scored over 90\% mean accuracy.
The x-axis is the ratio of number of programs that \name{} generates versus the number of programs that \nameold{} generates.
The y-value is the ratio of learning time of \name{} versus \nameold{}.
These ratios are acquired by dividing means, the mean of \name{} over that of \nameold{}.

Looking at x-axis values, of the 54 problems plotted all require fewer programs when run with \name{}.
Looking at the y-axis, the learning times of 51 problems are faster for \name{}.
\rolf{TODO: This analysis should be expanded.}

Overall, these results show that, compared to \nameold{}, \name{} typically needs fewer programs and less time to learn programs.
This suggests that the answer to questions \textbf{Q1} and \textbf{Q2} is yes.

\section{Conclusions}
\label{sec:conclusion}

We introduced a method for using fine-grained failure explanation to derive fine-grained hypothesis space constraints.
We illustrated this general method by a new SLD-based algorithm to identify failing sub-programs at the granularity of literals.
We introduced an ILP system with failure explanation, \name{}, and experimentally showed that enabling failure explanation can drastically reduce hypothesis space exploration and learning times.

\subsection{Limitations and future work}
Application of sub-program based failure explanation is not restricted to fully automated program synthesis.
For example, our SLD-based algorithm could be used for explainable AI purposes, e.g.~in interactive environments such as tutor systems which help teach Prolog.

While not documented here, our approach works without modification in combination with an extension of \popper{} which supports predicate invention \cite{poppi}.
In an orthogonal direction, ILP noise handling methods could leverage failure explanation, e.g.~by learning that the training error of a failing sub-program is as bad as the original program.

There are interesting theoretical questions to be worked out.
As seen in Experiment 2, it appears that many smaller sub-programs are key to effective pruning.
It should be possible to quantify the (theoretical) effectiveness of sub-program based pruning, e.g.~with respect to the size of a sub-program and hypothesis space parameters such as the number of predicates.
In general, future work should try to determine characteristics of problems that allow or preclude effective pruning based on failure explanation.

We require retesting of a sub-program derived from a hypothesis failing on a positive example to determine if this sub-program fails on the same example.
This retesting is especially costly if there are many sub-programs, as is more likely to happen for bigger programs. 
Theoretical work is needed to identify cases where it follows from the original SLD-tree only having failing branches that the SLD-tree for the sub-program has no successful branch either.
This would allow for eliding some of the expensive retesting that \name{} does.
\rolf{@AC: If incomplete (sub-)program gets recursive call added on to it is still incomplete. We currently do not do this pruning.}

\ac{not adding redundant constraints}

Another major avenue for future work is leveraging fine-grained failure explanation for learning programs from logic fragments extending beyond definite programs.
It should be possible to support negation-as-failure to a degree, e.g.~by saying that clauses defining a predicate that occurred negated in a hypothesis are also responsible for a failure.
Work on \emph{justifications} for Answer Set Programming \cite{ASPjustifications} could be used for fine-grained pruning whilst learning ASP programs.

Although we have shown that failure explanation can drastically reduce learning times, there is still much scope for improvement.
For instance, Experiment 2 had the following failing sub-program occur:
\[
\left\{
\begin{array}{l}
\emph{f(A,B) $\leftarrow$ element(A,C),head(A,D),odd(C),even(C)}
\end{array}
\right\}
\]
Straightforward reasoning tells us literal \emph{head(A,D)} is not relevant to the failure of this sub-program.
Furthermore, we should be able to lay the blame on just the last two literals.

\rolf{References still need to be cleaned up. (Probably shouldn't be problematic to just let reviewers complain about it and fix it then.)}
\bibliographystyle{plain}
\bibliography{00-bibliography}
\appendix
\newpage

\section{Experiment 2: Metagol Settings}
\label{app:metarules}

The following metarules were used for running Metagol in the programming puzzles experiment.

\begin{center}
\small
\centering
\begin{minipage}{.44\linewidth}%
\begin{lstlisting}[frame=single]
P(A):-Q(A).
P(A):-Q(A),R(A).
P(A):-Q(A,B),R(B).
P(A):-Q(A,B),P(B).
P(A):-Q(A,B),R(A,B).
\end{lstlisting}
\end{minipage}%
\hspace{2ex}
\begin{minipage}{.49\linewidth}%
\begin{lstlisting}[frame=single]
P(A,B):-Q(A,B).
P(A,B):-Q(A,B),R(A,B).
P(A,B):-Q(A),R(A,B).
P(A,B):-Q(A,B),R(B).
P(A,B):-Q(A,C),R(C,B).
P(A,B):-Q(A,C),P(C,B).
\end{lstlisting}
\end{minipage}%
\end{center}

\section{Experiment 2: Hypothesis Space Settings}
\label{sec:puzzlesettings}

The following hypothesis space settings were used in the programming puzzles experiment:

\begin{table}[ht]
\scriptsize
\centering
\begin{tabular}{c||c|c|c||c|c|c|c|c|c|c|c|c|c|c|c|c|c}
\textbf{problem}&
\rotatebox{90}{max \#clauses}&
\rotatebox{90}{max \#literals}&
\rotatebox{90}{max \#variables}&
\rotatebox{90}{sum/3}&
\rotatebox{90}{cons/3}&
\rotatebox{90}{snoc/3}&
\rotatebox{90}{head/2}&
\rotatebox{90}{tail/2}&
\rotatebox{90}{element/2}&
\rotatebox{90}{decrement/2}&
\rotatebox{90}{increment/2}&
\rotatebox{90}{qeg/2}&
\rotatebox{90}{even/1}&
\rotatebox{90}{odd/1}&
\rotatebox{90}{one/1}&
\rotatebox{90}{zero/1}&
\rotatebox{90}{empty/1}
\\
\hline
addhead/2&3&7&6&
x&x& &x&x& &x& &x&x&x&x&x&x\\
\hline
dropk/3&3&6&5&
x&x& &x&x& &x&x&x&x&x&x&x&x\\
\hline
droplast/2&3&6&5&
 &x& &x&x& &x&x&x&x&x&x&x&x\\
\hline
evens/1&2&6&5&
x& & &x&x& & & &x&x&x&x&x&x\\
\hline
finddup/2&2&6&5&
x& & &x&x&x&x& &x&x&x&x&x&x\\
\hline
last/2&3&7&6&
 & & &x&x& & & &x&x&x&x&x&x\\
\hline
len/2&2&6&6&
x& & &x&x& &x& &x&x&x&x&x&x\\
\hline
member/2&3&7&6&
x& & &x&x& &x& &x&x&x&x&x&x\\
\hline
odd1even2/2&3&6&5&
x& & &x&x& & & &x&x&x&x&x&x\\
\hline
reverse/2&3&5&5&
 & &x&x&x& &x& &x&x&x&x&x&x\\
\hline
sorted/1&3&6&5&
 & & &x&x& &x& &x&x&x&x&x&x\\
\hline
sumlist/2&2&6&5&
x&x& &x&x& &x& &x&x&x&x&x&x\\
\hline
threesame/1&3&7&6&
x& & &x&x& &x& &x&x&x&x&x&x\\
\end{tabular}
\end{table}

\end{document}